\RequirePackage{fix-cm}
\documentclass{svjour3}                     % onecolumn (standard format)
\smartqed  % flush right qed marks, e.g. at end of proof
\usepackage{times}
\usepackage{epsfig}
\usepackage{graphicx}
\usepackage{amsmath}
\usepackage{amssymb}
\usepackage{algorithm}
\usepackage{algorithmic, mathrsfs, bm}
\usepackage{subfigure}
\usepackage{marvosym}
\usepackage{multirow}
\usepackage[pagebackref=true,breaklinks=true,letterpaper=true,colorlinks,bookmarks=false,citecolor=blue]{hyperref}

\usepackage{url}
\journalname{Multimedia Tools and Applications}
\begin{document}

\title{Particle Filter Re-detection for Visual Tracking via Correlation Filters
}

\author{Di Yuan\textsuperscript{*} \and Xiaohuan Lu \and Donghao Li \and Yingyi Liang\textsuperscript{*} \and Xinming Zhang\textsuperscript{*}}


\institute{
        \textsuperscript{*} Corresponding authors.\\
         D. Yuan and X. Lu are contributed equally to this work and should be considered co-first authors.\\
        Di Yuan \at School of Computer Science and Technology, Harbin Institute of Technology, Shenzhen, China.
              \email{dyuanhit@gmail.com}.
           \and
           Xiaohuan Lu \at School of Computer Science and Technology, Harbin Institute of Technology, Shenzhen, China.
              \email{luxiaohuanok@126.com}.
              \and
           Donghao Li  \at School of Computer Science and Technology, Harbin Institute of Technology, Shenzhen, China.
                \email{lidh@hit.edu.cn}.
                \and
           Yingyi Liang \at School of Computer Science and Technology, Harbin Institute of Technology, Shenzhen, China.
                \email{liangyingyi002@foxmail.com}.
                \and
        Xinming Zhang \at School of Science, Harbin Institute of Technology, Shenzhen, China.
                \email{xinmingxueshu@hit.edu.cn}.
}

\date{Received: 05 March 2018 / Accepted: 23 October 2018}

\maketitle

\begin{abstract}
Most of the correlation filter based tracking algorithms can achieve good performance and maintain fast computational speed. However, in some complicated tracking scenes, there is a fatal defect that causes the object to be located inaccurately.
In order to address this problem, we propose a particle filter redetection based tracking approach for accurate object localization. During the tracking process, the kernelized correlation filter (KCF) based tracker locates the object by relying on the maximum response value of the response map; when the response map becomes ambiguous, the KCF tracking result becomes unreliable. Our method can provide more candidates by particle resampling to detect the object accordingly.
Additionally, we give a new object scale evaluation mechanism, which merely considers the differences between the maximum response values in consecutive frames.
Extensive experiments on OTB2013 and OTB2015 datasets demonstrate that the proposed tracker performs favorably in relation to the state-of-the-art methods.
\end{abstract}

\section{Introduction}
Visual object tracking is one important topic in computer vision and plays a necessary role in numerous applications, such as video surveillance, automobile navigation, human-computer interfaces, robotics and driverless vehicle.
Although substantial progress has been proposed in recent years, achieving higher efficiency with lower computation complexity in visual object tracking is still a tough problem to solve.

Many different methods have been proposed for visual object tracking in succession in recent decades\cite{KAP:Isard,IEEE:Hossain,MOT,Zhang2017Robust,Li2014A}.
Filtering technology has made great progress in the field of image processing\cite{He20103-D,Fazli2009Particle,He2010Writer,Mai2016Optimization,He2009Texture,He2008Writer}.
Considering that particle filters assume non-linearity and non-Gaussianity assumption
to estimate problems and have high performance\cite{KAP:Isard},
Kabir and Chi-Woo\cite{IEEE:Hossain} develop a observation model based on the robustness of phase correlation in a particle filter framework for visual object tracking to address occlusion.
Li $et$ $al.$\cite{IEEE:Li} propose a visual object tracking method based on adaptive background modelling to improve the robustness of the particle filter framework.
There are also some improved trackers that have more precision and robustness than the traditional particle filter based trackers\cite{Yi2015Single,Ou2017Object,Yu2010Infrared}.
Although particle filter based trackers have some advantages, the high computational complexity remains a fatal flaw.
In this paper, we use a particle resampling strategy to provide more target candidates and use the correlation filter to choose the best one as the target object to improve the computational efficiency.

Recently, correlation filter based tracking algorithms have achieved remarkable results\cite{IEEE:MOSSE,He2015One,Springer:DSST,He2016Robust,liu2017deep}.
Typically, the design of the correlation filter usually places the peak of the response in the scene as the tracking target, and place the low response position as the background.
Although the filter can locate the tracking target effectively, the training process requires a large number of samples, which reduces the tracking speed.
By using adaptive training schemes, Bolme $et$ $al.$\cite{IEEE:MOSSE} propose a minimum output sum of squared error (MOSSE) filter whose tracking results are quite robust and effective.
After that, a series of correlation filter based tracker were developed\cite{Springer:SAMF,Springer:RPAC,fDSST,Multi-view}.
Although these correlation filter based trackers have achieved some pretty better performances in visual object tracking, all of them  have the limitation of being excessively dependent on the maximum response value.
When the response map become unreliable, the maximum response value becomes smaller. Under these circumstances, the object is determined by the response map may drift or become lost, so an efficient redetection mechanism is very important in the tracking algorithm\cite{WEN2008TWO,Yi2017Unified,Jing2015Super,Lai2016Approximate,Shi2016Two,Qi2017Structure,yi2017joint}.

To provide a more credible candidate for the object target, we develop an effective redetection model for visual tracking.
In each frame after initialization, an image patch of a previous estimated position is cropped as the search window input and the HOG features are extracted to better describe it.
Subsequently, convolution between input features and the correlation filter is performed in the frequency domain.
After that, a response map is obtained by inverse fast Fourier transform (IFFT) and we can get the maximum response value from it.
If the maximum response value is larger than the threshold, the coordinate of the maximum response value is taken as the object new position.
If the maximum response value is less than the threshold, the location of the target is redetected by using the particle filter to resample more candidates.
Lastly, extracted the appearance in the newly estimated position is extracted to train and update the correlation filter.

The main contributions of this paper can be summarized as follows:
\begin{itemize}
\item We propose an efficient method for accurately locating the  tracking object by particle filter redetection. This method allows us to redetect the location of  target object if the result of the correlation filter tracking  is ambiguous or unreliable.
\item A novel scale-evaluation strategy is given by comparing the relationship of the maximum response values in consecutive frames.
    This scale evaluation mechanism can effectively reduce the impact of variations in the scale of the target on the performance and increase the robustness of the algorithm.
\end{itemize}

The rest of this paper is structured as follows. We first introduce some related works in Section \ref{Rw}.
Next, we propose the particle filter redetection tracker via correlation filters, including the introduction to the basic KCF tracker, the particle filter redetection model, the scale evaluation and the model update in Section \ref{OurM}.
Then, we present the implementation details of our tracker in Section \ref{ImD}.
Subsequently, we introduce the evaluation criterion and evaluate our approach on comprehensive benchmark datasets in Section \ref{Exp}.
Finally, we briefly present the conclusion of our work in Section \ref{Conc}.

\section{Related Works} \label{Rw}
Visual object tracking has been studied extensively and have multifarious applications in the real world scenes.
As an comprehensive review on particle filter technology and correlation filter technology is not necessary for this paper, we have just reviewed the works related to our method for simplicity, which include the particle filter based trackers and the correlation filter based trackers.

\subsection{Particle Filter Based Trackers}\label{PFT}
Particle filter based algorithms have been studied in visual object tracking for many years and their variations  are still widely used nowadays\cite{IEEE:Hossain,IEEE:Li,Yuan2017Patch,Mai2016Optimization,Fazli2009Particle}.
The traditional particle filter algorithm implements a recursive Bayesian framework by using the nonparametric Monte Carlo sampling method, which can effectively track the target objects in most scenes\cite{KAP:Isard}.
However, due to the limitation of initializing the particle number and the target template artificially, the number of particles is hard to decide and the target template selection is not accurate enough.
Li $et$ $al.$\cite{IEEE:Li} presented an improved particle filter algorithm that achieved semi-automatic initialization of the tracking object.
In view of the particle filter framework, Kabir and Chi-Woo\cite{IEEE:Hossain} proposed a phase dependent robust observation model and introduced an optimization method to improve the precision.
Because particle filters are able to model the uncertainty of object movements, which can provide a robust tracking framework,
it can also consider multiple state hypotheses simultaneously.
Zhou $et$ $al.$\cite{Zhou2016Adaptive} integrated multiple cues into a particle filter framework and introduced a quality function to calculate the reliability of each cue.
Mai $et$ $al.$\cite{Mai2016Optimization} developed a particle filter based tracker by using color distributions features and  utilized the computing power of embedded systems to reduce the complexity of the tracker.
By embedding deterministic linear prediction in stochastic diffusion, an adaptive method has been proposed to adjust the number of samples according to an adaptive noise component \cite{Zhou2004Visual}.
Although these particle filter frameworks have achieved some good performances, they still suffer from one drawback: high computational complexity.

Unlike these methods, our method uses the property of a circulant matrix and the conversion between time and frequency domains to reduce the computational cost and can handle situations, in which the target object becomes lost by the correlation filter based tracker since the particle filters used by our method can produce high accuracy predictions from previous observations by the particle resampling strategy.

\subsection{Correlation Filter Based Trackers}\label{CFT}
Since correlation filter operation can convert the convolution operation of two image blocks into Fourier domain element-wise products, it has been applied to visual object tracking thanks to the fast computational speed.
Bolme $et$ $al.$ \cite{IEEE:MOSSE} introduced an adaptive correlation filter by minimizing the output sum of the squared error to make the tracking strategy simpler and more effective.
In an evaluation of online visual tracking approaches,  Henriques $et$ $al.$\cite{Springer:CSK} proposed a CSK tracker that can provide good performance and a high calculation speed.
These two trackers both use the single-channel gray value feature.
Danelljan $et$ $al.$\cite{IEEE:CN} improved the CSK methods by using the color attributes.
In \cite{IEEE:KCF}, the KCF method further improves the efficiency of the CSK tracker by using HOG features and the kernel method to transform the non-linear regression problem into a linear regression.
For the scale evaluation problem, the DSST\cite{Springer:DSST}(discriminative scale space tracking) tracker uses the HOG feature to learn an adaptive multi-scale correlation filter to handle the scale change of the object target.
Zhang $et$ $al.$ \cite{Zhang2016In}, exploited the circulant structure property of a target template to improve sparse representation based trackers.
For improve the robustness of the algorithm, some local patches or parts based correlation filter trackers  have also been developed\cite{Ou2016Multi,Li2015Reliable,Guo2015Robust,Liu2015Real,Liu2016Structural}.
Li $et$ $al.$ \cite{Li2015Reliable} introduced reliable patches, whose distributions are under a sequential Monte Carlo framework, to exploit the use of local contexts to carry out the tracking task.
In\cite{Liu2015Real}, a part based multiple correlation filter is proposed to preserve the structure of the object target by adopting a Bayesian inference framework and a structural constraint mask to make the tracker robust.
Liu $et$ $al.$\cite{Liu2016Structural} proposed a part based structural correlation filter and exploited circular shifts of those parts to preserve the structure of the object target for visual tracking.
However, these correlation filter based tracking methods are exceedingly dependent on the maximum response value. Therefore, these methods may lose their tracking object when the maximum response value becomes ambiguous or unreliable.

Unlike the existing correlation filter based tracking methods, which are excessively dependent on the maximum response value to locate the target, we propose a particle filter redetection correlation filter tracker. When the correlation filter based tracking result become unreliable, the particle filter redetection method can exploit the particle resampling strategy to provide more object candidates, which can greatly enhance the robustness of the tracking method.

\section{The Proposed Method} \label{OurM}
In this section, we give the overall algorithm framework in Section \ref{Overview}, introduce the basic framework of kernelized correlation filter based tracker in Section \ref{CFTF}, propose the method of particle filter redetection for visual tracking in Section \ref{PFR}, give a simple but effective scale evaluation algorithm in Section \ref{Se}, and finally, propose the a model update strategy in Section \ref{ModUp}.

\subsection{Overview of the Proposed Method}\label{Overview}
\begin{figure}[!t]
\centering
\centerline{{\includegraphics[width=0.98\textwidth]{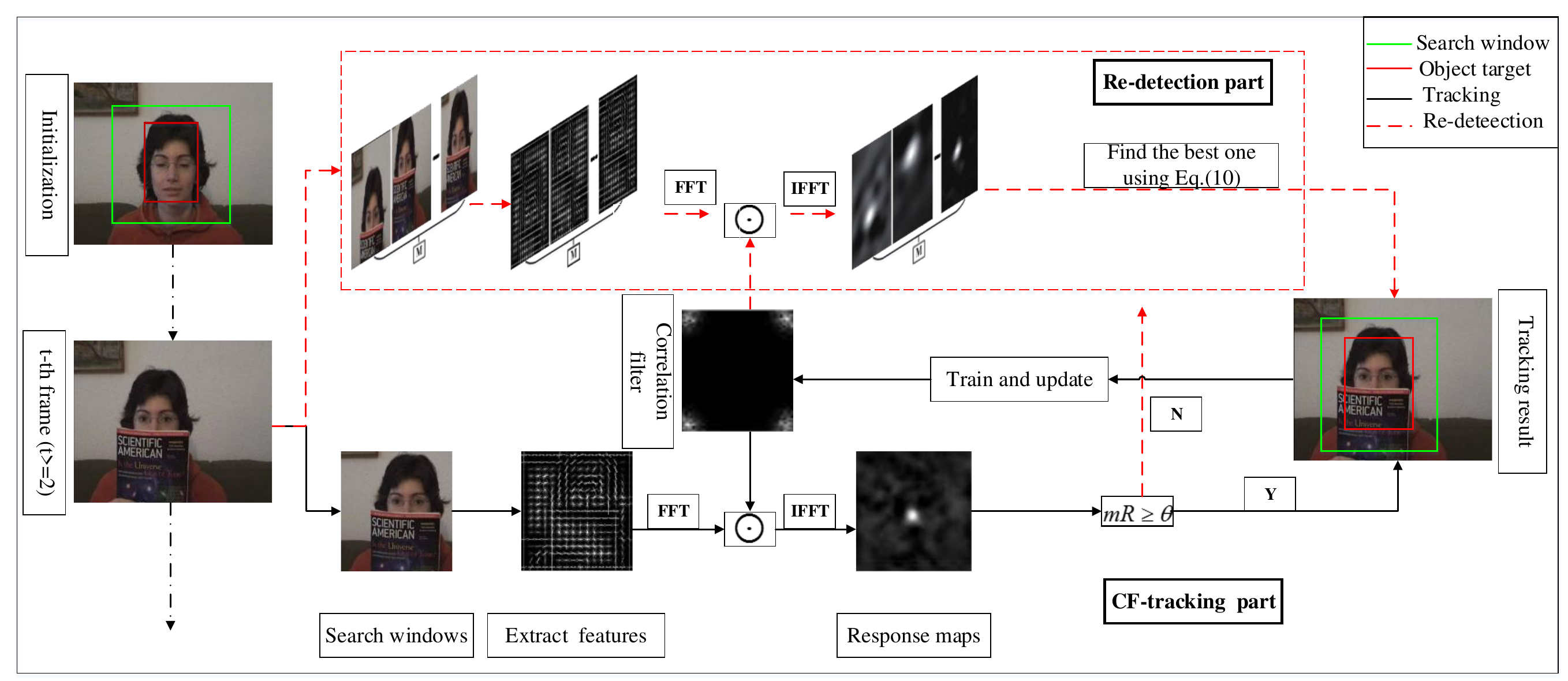}}}
\caption{The framework of our approach is comprised of two parts: the CF-tracking part and the redetection part. For the $t$-th frame, the maximum response value and the threshold are compared and the part that is to be used to track the target object is determined.
}
\label{fig:1}
\end{figure}

As illustrated in Figure \ref{fig:1}, the proposed approach consists of two parts: the CF-tracking part, which is used to track the target object directly, and the redetection part, which is used to re-detect the object target.
During the tracking process, the feature is extracted according to the known target position in the first frame, and the correlation filter is trained directly.
The  target object size of the first two frames never has no obvious changes, so we use the same object size in the first two frames.
Next, in $t$-th frame ($t \geq 2$),  the feature is extracted from the search window and the response map is computed through the known correlation filter.
Then, by comparing the maximum response value ($mR$) and the threshold ($\theta$), we can determine which part can be used to track the target object.
If $mR \ge \theta$, the tracker gives the tracking result directly; otherwise, the redetection part is used to track the target object.
Finally, when we get the tracking result, the result is used to train and update the correlation filter correspondingly until the last tracking frame.

\subsection{Kernelized Correlation Filter Based Tracking Framework}\label{CFTF}
Before the detailed discussion of our proposed framework and for completeness, we first revisit the details of the conventional KCF\cite{IEEE:KCF} based tracking method.
The KCF tracking method trains a classifier through dense sampling from an image patch.
Using kernel trick, the samples data matrix can have highly structured, and is thus allow to operation with cyclic shifts.
Besides, according to the Convolution Theorem, we can get the convolution of two patches in the spatial domain simply by using the element-wise product in the Fourier domain.
Therefore, for correlation filter based trackers, the computational efficiency can be greatly improved by Fast Fourier Transform (FFT) and its inverse transformation.
However, the KCF tracker uses the target object appearance to train and update its models, and if the object is heavily occluded or in fast motion, the tracker may fail to detect it.

The KCF tracker models the appearance of the target object by using a filter $w$, which is trained on an image patch $x$ of $M\times N$ pixels with HOG features. All the circular shifts of $x_{m,n}, where \ $(m,n)$ \ \in \{0,1,...M-1\}\times \{0,1,...,N-1\}$ are generated as the training samples for the filter with the Gaussian function label $y_{m,n}$. The filter $w$ can be acquired by minimizing the error between the training sample $x_{m,n}$ and the regression target $y_{m,n}$. The minimization problem is:

\begin{equation}\label{eq:kcf1}
w = \arg\min_w {\sum_{m,n}{|\left \langle \phi(x_{m,n}),w\right \rangle-y(m,n)|^2 + \lambda_1 \lVert w \rVert^2  }},
\end{equation}
where $\phi$ represents the mapping to a kernel space in the Hilbert space, $\left \langle. \right\rangle$ denotes the inner product, and $\lambda$ is a regularization parameter ($\lambda\geq 0 $). Since the label $y_{m,n}$ is not binary, the filter $w$ learned from the training samples contains the coefficients of a Gaussian ridge regression.

Using FFT to compute this problem, this objective function can be identically expressed as $w = \sum_{m,n} \alpha_(m,n)\phi(x_{m,n})$, so the solution of Eq.~(\ref{eq:kcf1}) can be acquired by the following formula:

\begin{equation}\label{eq:kcf2}
\alpha=\mathcal{F}^{-1}\left(\frac{\mathcal{F}(y)}{\mathcal{F}(k^{x})+\lambda}\right),
\end{equation}
where $\mathcal{F}$ and $\mathcal{F}^{-1}$ denote FFT and  IFFT, respectively.
The kernel correlation $k^{x}=\kappa (x_{m,n},x)$ is computed by the Gaussian kernel in the Fourier domain.
The vector $\alpha$ contains all the $\alpha_{m,n}$ coefficients.
The KCF model consists of the target object appearance $\hat{x}$ and the coefficients $\mathcal{F}(\alpha)$.

In the tracking process, a patch $z$ with the same size as $x$ is cropped from the new frame image.
The response score is calculated by:

\begin{equation}\label{eq:kcf3}
f(z)=\mathcal{F}^{-1}(\mathcal{F}(k^{z}) \odot\mathcal{F}(\alpha)),
\end{equation}
where $\odot$ denotes the element-wise product, $k^{z}=\kappa(z_{m,n},\hat{x})$, and $\hat{x}$ is the learned target object appearance.

\subsection{Particle Filter Re-detection Model}\label{PFR}
The fast motion, occlusion or background clutter of the target object can have a big impact on the tracking performance. For example, if there is a lot of background clutter, the KCF tracker maybe lose the target object because it is overly dependent on the maximum response value. Therefore, we propose a framework for the KCF tracker that can provide more target object candidates in the redetection part (Figure \ref{fig:1}, Re-detection part).

A particle filter is an efficient method of providing more reasonable target object candidates by using the particle resampling strategy.
The central idea is to use a set of random particles with related weights to represent posterior densities and estimate the values based on these samples and the related weights\cite{IEEE:Arulampalam,Lai2014Multilinear}.
It is based on the theory of the sequential Monte Carlo importance sampling method.
Suppose $s_{t}$ and $y_{t}$ are the state and the observation variables at time $t$ respectively. Mathematically, object tracking is based on the observations up to the previous time $t-1$ to find the most probable state at time $t$:

\begin{equation}\label{eq:pft1}
\begin{split}
s^t &=argmax p(s^t|y^{1:t}) \\
     &=argmax\int{p(s^t|s^{t-1})p(s^{t-1}|y^{1:t-1})ds^{t-1}}.
\end{split}
\end{equation}

The posterior distribution of the state variable is updated according to Bayes rule using the new observation  $y_{t}$ at time $t$:

\begin{equation}\label{eq:pft2}
 p(s^t|y^{1:t})=\frac{p(y^t|s^{t})p(s^{t}|y^{1:t-1})}{p(y^{t}|y^{1:t-1})}.
\end{equation}

The particle filter approach approximates the posterior state distribution $p(s^{t}|y^{1:t})$  by $n$ samples, which are called particles  $\{{s^t_{i}}\}^n_{i=1}$  with corresponding importance weights $\{{w^t_{i}}\}^n_{i=1}$ and the $w^t_{i}$ sum to 1.
The particles obey an importance distribution  $q(s^{t}|s^{1:t-1},y^{1:t})$ and the weights are updated as:

\begin{equation}\label{eq:pft3}
 w^t_{i}= w^{t-1}_{i} * \frac{p(y^t|s^{t}_{i})p(s^{t}_{i}|s^{t-1}_{i})}{q(s^{t}|s^{1:t-1},y^{1:t})}.
\end{equation}

When the state transition is independent of the observation, $p(s^{t}|s^{1:t-1},y^{1:t})$ is always simplified to a first-order Markov process  $q(s^{t}|s^{t-1})$. Meanwhile, the weights are updated as:

\begin{equation}\label{eq:pft4}
w^t_{i}= w^{t-1}_{i}p(y^t|s^t_{i}).
\end{equation}

For every frame, the tracker always uses the the particle with the largest weight as the tracking result.
Correspondingly, as in our tracking framework, we can use the particle with the largest maximum response value as the tracking result.

In our method, we think the correlation filter based tracker (CFT) is strongly dependent on the maximum response value of the response map. This situation may cause the tracker to lose its target object when the response map becomes ambiguous or unreliable. In order to ensure that our algorithm achieves a high performance, a particle filter redetection tracker (PFT) mechanism has been adopted. The main equations are as follow:

\begin{equation}\label{eq:dis}
\begin{cases}
    maxR \geq \theta \ \ \ \ \  tracking\ by\ CFT,  \\
    maxR <   \theta \ \ \ \ \ tracking\ by\ PFT ,    \\
  \end{cases}
\end{equation}
where $maxR$ denotes the maximum response value of the response map, which is obtained from the correlation filter based tracker, and $\theta$ is a threshold to determine whether the response map is credible or not.

For the particle filter part of our method, we utilize the advantage of the particle resampling mechanism to provide more reasonable object candidates.
In this redetection part, the number of image patches is set to $M$ with the same size of search window, and the image patches obey the normal distribution centered on the position of the previous target object.
Given the object appearance model $\hat{x}$ and the coefficients $\mathcal{F}(\alpha)$, each particle image can be guided toward the modes of the target state distribution by using its circular shifts. For each image patch (also called a particle) $m\in {1,2,...,M}$, the HOG features are extracted and a correlation calculation is performed between the HOG features and the correlation filter in the frequency domain based on the Convolution Theorem.
After that, IFFT is used to obtain the spatial response map. This is expressed in the mathematical model as:

\begin{equation}\label{eq:cfpft1}
R_m=\mathcal{F}^{-1}(\mathcal{F}(<z_m,\hat{x}>) \odot\mathcal{F}(\alpha)),
\end{equation}
where $z_m$ denotes the $m$-th particle corresponding to the image patch and $R_m$ denotes the corresponding response map.

We can choose the image patch with the best maximum response value as the center of the target object, because mathematically,

\begin{equation}\label{eq:tr}
    maxR_{pf} = max{\{maxR_1,maxR_2,...,maxR_{M}\}},
\end{equation}
where $maxR_{pf}$ denotes the best particle, which corresponds to the one with the maximum response value, and $maxR_m$ denotes the $m$-th particle corresponding maximum response value.

In this approach, we use the particle filter to choose more search windows when the response map given by a single search window is ambiguous or unreliable. In this way, we can find more target object  candidates, which can make the tracking results more robust and efficient.

\subsection{Scale Evaluation}\label{Se}
We can only obtain the object center position through the maximum value of the response map in correlation filter tracking framework, but there is no scale estimation of the tracking object\cite{IEEE:KCF,Springer:CSK}.
However, scale variation is one common challenging aspect and can influence the accuracy and performance in visual tracking process\cite{Springer:DSST}.
In this section, we give a simple but effective mechanism for detection of scale changes depend on the relationship between the maximum response values of the consecutive frames.

For most tracking methods, the model or template size of the object is fixed as either a manually set or a initial object size\cite{You2014A,Zhang2015Robust,ERT}.
In order to handle candidate images of different sizes, the candidate images patch are usually adjusted to the same size by the affine transformation model.
But the affine transformation model has more parameters, which can lead to high computational cost and reduce the tracking efficiency.
The rate of change of the maximum response between consecutive frames is negatively related to the change of the object size, because the response map of the object obeys a normal distribution.
Therefore, we use the change rate between the maximum responses of consecutive frames to determine the change of object size.
In tracking image sequences, due to the size of object is gradually changing and is accompanied by a certain degree of attenuation effect, for simplicity, we only consider the changing trend of the object size instead of the accurate values, so we merely consider whether the size of the object becomes smaller or larger or remains unchanged.

For the correlation filter based tracker, the initial target size $size_1=(h_1,w_1)$ has been given.
We set the object size as the initial size for the second frame.
Then, if we know the object size of the ($t-1$)-th frame (where $t>2$), we can determine the trend of the change in the object size in the $t$-th frame. The direction $d_t$ for the $t$-th frame can be determined as:

\begin{equation}\label{eq:se1}
 d_t= \frac{maxR_t}{maxR_{t-1}}-\frac{maxR_{t-1}}{maxR_{t-2}},
\end{equation}
where $maxR_t$ denotes the maximum response value of the $t$-th frame; if $d_t>\phi$ the target size of $t$-th frame becomes smaller; if $d_t<\psi$ the target size of $t$-th frame becomes larger; otherwise the target size of the $t$-th frame does not change. $\phi$ and $\psi$ are two thresholds that are used to determine the direction of the size of the target object.

The target size of the $t$-th frame can be calculated as:

\begin{equation}\label{eq:se2}
 size_t= size_{t-1}*s_t,
\end{equation}
where $size_t$ denotes the target size of the $t$-th frame and $s_t$ denotes the scale factor of frame $t$, which determined by $d_t$. If $d_t>\phi$ the scale factor $s_t=0.98$; If $d_t<\psi$  the scale factor $s_t=1.02$; otherwise  the scale factor $s_t=1$.
 We can use Eq.(~\ref{eq:se1}) to obtain the direction of the object scale change and use Eq.(~\ref{eq:se2}) to achieve the optimal size of the object in frame $t$.

\subsection{Model Update}\label{ModUp}
Model updating is an important step in visual tracking.
In the process of tracking, the object appearance often changes with the factors of rotation, scale and posture.
Therefore, the filter needs to be quickly updated to accommodate the changes in the object tracking process.
In this paper, we adopt a linear update model to update the filter\cite{IEEE:MOSSE,Qian2011Accurate}.
The method only exploits the current frame target $x_t$ to update the filter:

\begin{equation}\label{eq:mu1}
\begin{aligned}
& H_{t} = \frac{y}{\hat{x}_t*\hat{x}_t{^*}+\lambda} \odot\hat{x}_t,\\
& W_t = (1 - \gamma )W_{t-1} + \gamma H_{t},\\
\end{aligned}
\end{equation}
where $W_t$ denotes the updated correlation filter model of the target in the $t$-th frame, $H_t$ denotes the  correlation filter of the $t$-th frame, and $\gamma$ is the learning rate, which is used to update the correlation filter in the current frame.

\section{Implementation Details}\label{ImD}
In this section, we first represent the overall tracking process of our proposed tracking method, and then describe the parameter settings of our experiments.
The integral framework of our approach is given in Algorithm ~\ref{alg:1}.
Our tracker begins with the object position in the first frame image, and we can use it to train a correlation filter.
In the next frame image, we can extract the HOG features from a search window and convolve with the correlation filter to obtain a response map.
The maximum response value is compared with a threshold to determine which method should be used to find the new position of the target object.
If the maximum response value is more than the threshold, the position is found directly by KCF tracking; otherwise, the location of the target is detected by using a particle filter to resample more candidates. Then, the appearance in the newly estimated position is extracted for training and updating of the correlation filter.
The whole process is repeated until the position of the target object is given in the last frame image.

Because the size of the target object has no obvious change in the first two frames, we use the same size for the first two frames and set the number of particles $M=100$, the judgement threshold $\theta = 0.05$, and the direction thresholds $\phi = 0.1$ and $\psi = -0.1$.
The search window size is set as $sz$-$window$ $=2*sz$, that is, twice the target object size.
The regularization parameter $\lambda$ of the CF model is set as $0.01$. In the part of scale evaluation, the three scale weights change directions $s_t$ are set as $\{0.98,1,1.02\}$ and the default setting is set for the basic KCF tracker\cite{IEEE:KCF}.

\begin{algorithm}[h]
 \caption{ Correlation filter based particle filter redetection framework (CFPFT)}
  \label{alg:1}
  \begin{algorithmic}[1]
  \STATE \textbf{Inputs}: the initial target bounding box $b_1$ , the target size $sz$, the search window  $sz$-$window$, the initial tracking frame $I_1$ , the model learning rate $\gamma$, the particle number $M$, the judgment threshold $\theta$, and the direction thresholds $\phi$ and $\psi$ .
  \STATE \textbf{Outputs}: The position and scale of the target in each frame.
  \STATE Extract the target features  from $I_1$ with area $b_1$;
  \STATE Train the initial models $mod_1$  with Eq.~(\ref{eq:kcf2}) and $b_1$;
 \IF{$t < T$, where $t$ is the number of the current frame and $T$ is the total number of tracking frames,}
 \STATE Evaluate the scale change and get the optimal scale factor $d_t$ with Eqs.~(\ref{eq:se1}) and (\ref{eq:se2});
 \STATE Crop the search window with $sz$-$window$ $*$ $d_t$ from the current frame $I_t$ and extract the features from the search window;
 \STATE Compute the correlation filter response with Eq.~(\ref{eq:kcf3});
 \STATE For the image in the $t$-th frame (where $t>2$), determine the difference between the maximum response value $maxR$ and the threshold value $\theta$ with Eq.~(\ref{eq:dis}).
 \IF{ $maxR$ $\geq$ $\theta$ }
 \STATE Get the target position of the current frame $t$  and the target object size is $sz_t=sz_{t-1}$ $*$ $s_t$;
 \ENDIF
 \IF{ $maxR$ $<$ $\theta$ }
 \STATE Get $M$ search windows from current frame and calculate the corresponding response map with  Eq.~(\ref{eq:cfpft1}).
 \STATE Choose the best candidate as the correlation filter response map with Eq.~(\ref{eq:tr});
 \STATE Get the object position of the current frame $t$, and the target object size is $sz_t = sz_{t-1}$;
 \ENDIF
 \STATE Get the correlation filter model $W_{t}$ with the current target object and update it with Eq.~(\ref{eq:mu1});
 \ENDIF
  \end{algorithmic}
\end{algorithm}

\section{Experiments}\label{Exp}
We evaluate our proposed method on the datasets OTB2013 and OTB2015 \cite{IEEE:OTB2013,Wu2015Object}.  The dataset OTB2013 has $51$ different sequences and categorizes these sequences with 11 attributes, namely, fast motion (FM), background cluster (BC), motion blur (MB), deformation (DEF), illumination variation (IV), in-plane rotation (IR), low resolution (LR), occlusion (OCC), out-of-plane rotation (OR), out of view (OOV) and scale variation (SV). The dataset OTB2015 includes $100$ different sequences.
Our method is implemented in MATLAB and run at round $22$ frames per second on a PC with an Intel Core-$i3$-$4170$ CPU ($3.70$ GHz) and $8$ GB of RAM.

\subsection{Evaluation Criterion}\label{ImSet}
In order to evaluate the performance of our proposed algorithm, we use three classes of evaluation indexes proposed in OTB2013: One-Pass Evaluation (OPE), Temporal Robustness Evaluation (TRE), and Spatial Robustness Evaluation (SRE). OPE is a traditional evaluation method that runs trackers on each sequence just once.
For TRE, each compared tracking method is evaluated numerous times from different starting frames across a video sequence. Each tracker is evaluated from a particular starting frame, with the initialization of the corresponding ground-truth object state in each evaluation, and the experiments are implemented $20$ times with different starting frames in every video sequence.
The SRE evaluation generates the object states by  shifting or scaling the ground-truth bounding box of an object slightly, and the experiments are implemented $12$ times with different spatial perturbations.
With TRE and SRE, the robustness of each evaluated trackers can be comprehensively interpreted.

After running the trackers, precision plots and success plots are applied to the present results.
Precision plots show the percentages of frames whose estimated locations lie within a given threshold distance from
the ground-truth centers.
With regard to the success plots, an average overlap measure is the most appropriate for tracker comparison\cite{IEEE:Cehovin}, as it accounts for both size and position. For this purpose, we use the typical criterion of the Pascal VOC Overlap Ratio (VOR)\cite{Springer:Everingham}.
Given the bounding $B_R$ of the result and the bounding box $B_G$ of the ground truth, the VOR can be computed as:

\begin{equation}\label{eq:ec1}
VOR = \frac{Area\{B_R \cap B_G \}}{Area\{B_R \cup B_G \}},
\end{equation}
where $\cap$ and $\cup$ denote the intersection and union of two regions, respectively.
Afterwards, a frame whose VOR is larger than a threshold is termed a successful frame, and the ratios of successful frames at the thresholds ranged ranging from $0$ to $1$ are plotted in the success plots.

\subsection{Experimental Evaluation}\label{Eotb}
In this section, we show the experimental results for the OTB2013\cite{IEEE:OTB2013} and OTB2015\cite{Wu2015Object}.
For these visual tracking benchmarks, the experimental results are illustrated by precision plot (or rate) and success plot (or rate).
The precision plot shows the percentage of successfully tracked frames in the whole sequence and evaluates the  performance of the algorithms with Center Location Error (CEL) in pixels, which ranks the trackers as the precision score at $20$ pixels.
The success plot shows the percentage of successfully tracked frames using the VOR threshold ($0.5$ is usually taken as the threshold), while the Area Under the Curve (AUC) is used as the metric for ranking.

\subsubsection{Evaluation with OTB2013}
In this section, we analyze our approach on the OTB2013\cite{IEEE:OTB2013} benchmark by demonstrating the impact of our contributions.
For the OTB2013\cite{IEEE:OTB2013} benchmark, the performances of all the tracking methods are measured by the OPE, TRE, and SRE mechanisms.

We compare our method with $31$ representative algorithms, which include $29$ algorithms given in the OTB2013 benchmark and two representative algorithms based on the correlation filter, namely, KCF\cite{IEEE:KCF} and DSST\cite{Springer:DSST}.

\begin{figure*}[!t]
\centering
\subfigure[The precision plots of OPE]{\includegraphics[width=0.48\textwidth]{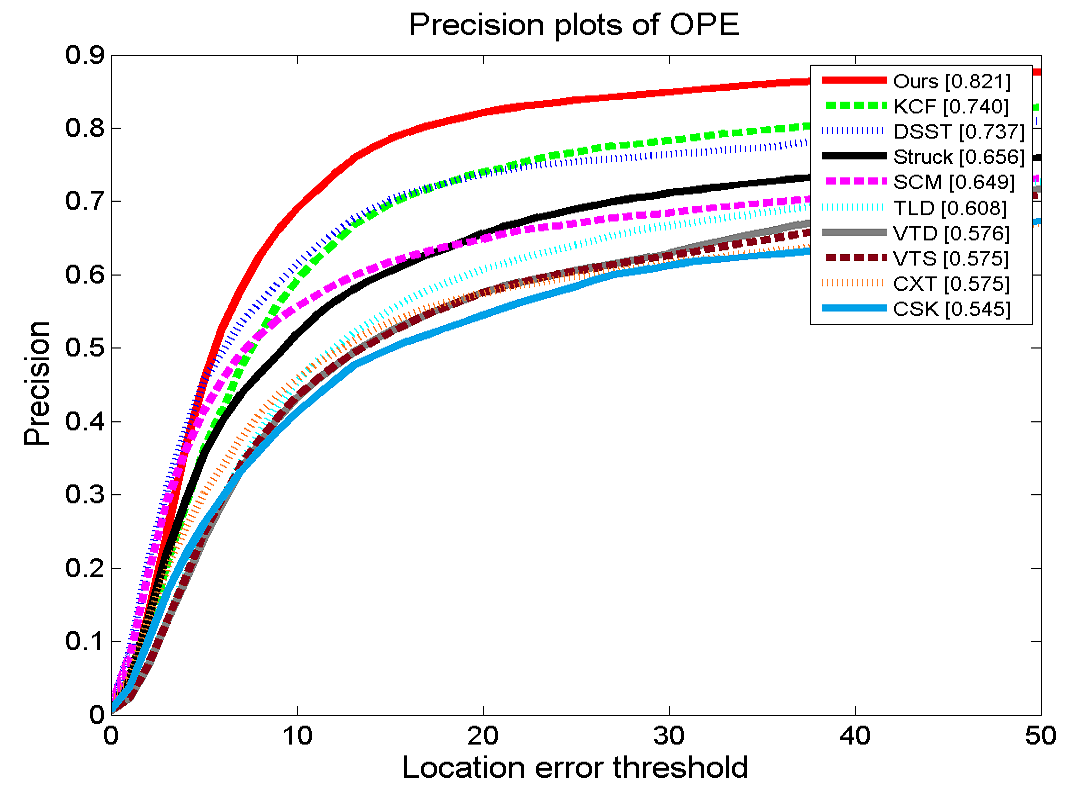}}
\subfigure[The success plots of OPE]{\includegraphics[width=0.48\textwidth]{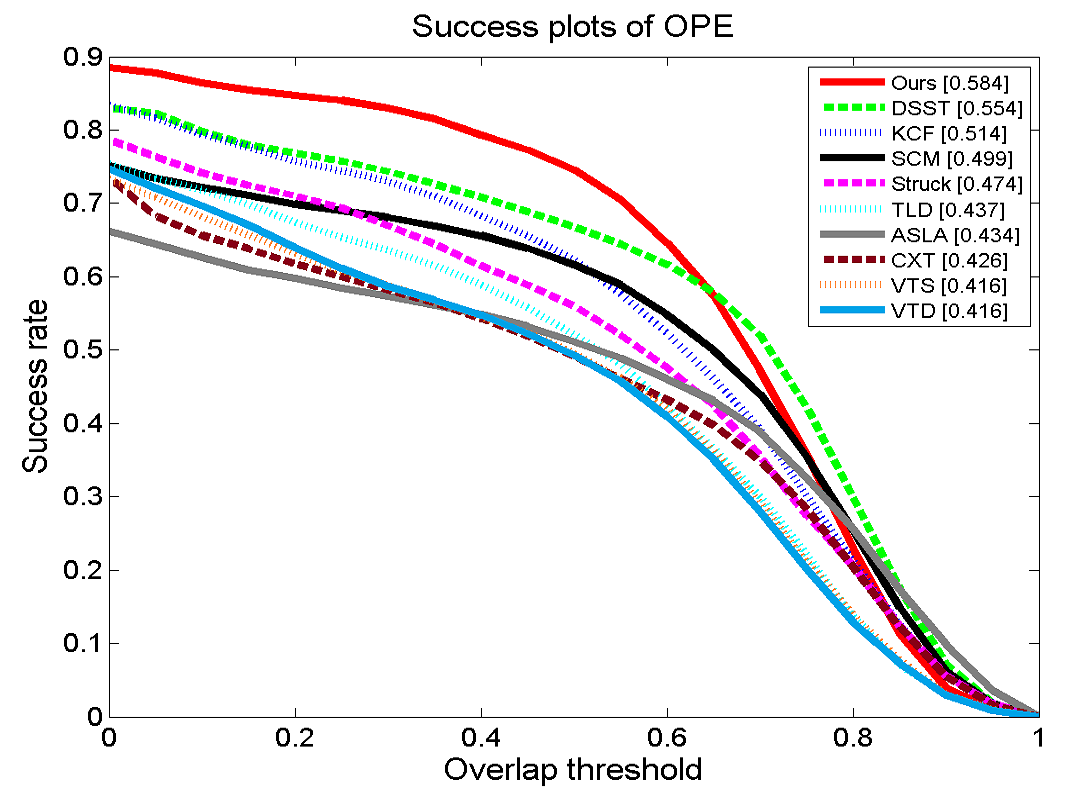}}
\caption{
Precision and success plots of OPE on the OTB2013 benchmark.
 The numbers in the legend indicate the representative precision at $20$ pixels for precision plots and the average area-under-the-curve scores for success plots.
}
\label{fig:ope}
\end{figure*}

To make the results clear, we only plot the top 10 ranked trackers in the precision and success plots. As shown in Figure~\ref{fig:ope}, our proposed CFPFT tracker achieves top rank and the best performance with a large margin in all the tracking plots.
Specifically, the proposed tracker achieves a ranking score $0.584$ for the success plot and a ranking score $0.821$. Compared with the KCF tracker, which has a  success ranking score $0.514$ and a precision ranking score $0.737$, our CFPFT tracker has obtained improvements over $13.62\%$ and $10.95\%$, respectively. Even compared with  DSST, which has a success ranking score $0.554$  and a precision ranking score $0.737$, our tracker also has obtained improvements over $5.42\%$ and $11.40\%$, respectively. This demonstrates that the idea of the redetection mechanism for tracking is effective and promising in practice.

\begin{figure*}[!t]
%\centering
\subfigure[Fast motion]{\includegraphics[width=0.245\textwidth]{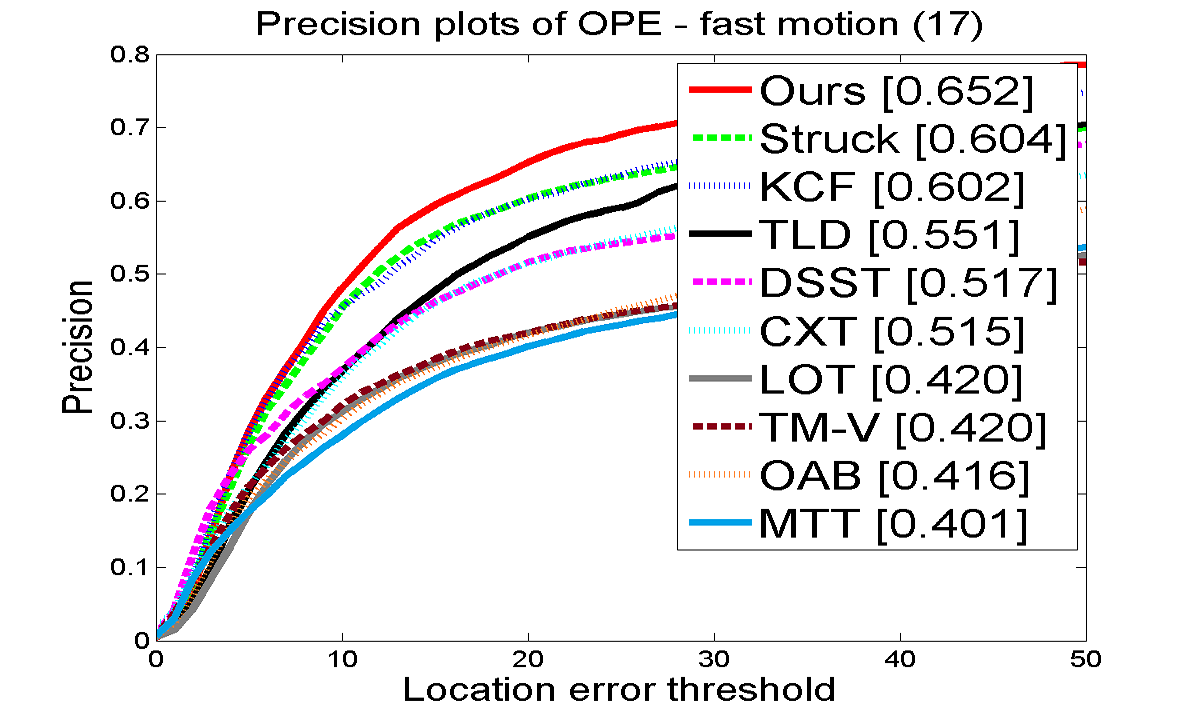}}
\subfigure[Background clutter]{\includegraphics[width=0.245\textwidth]{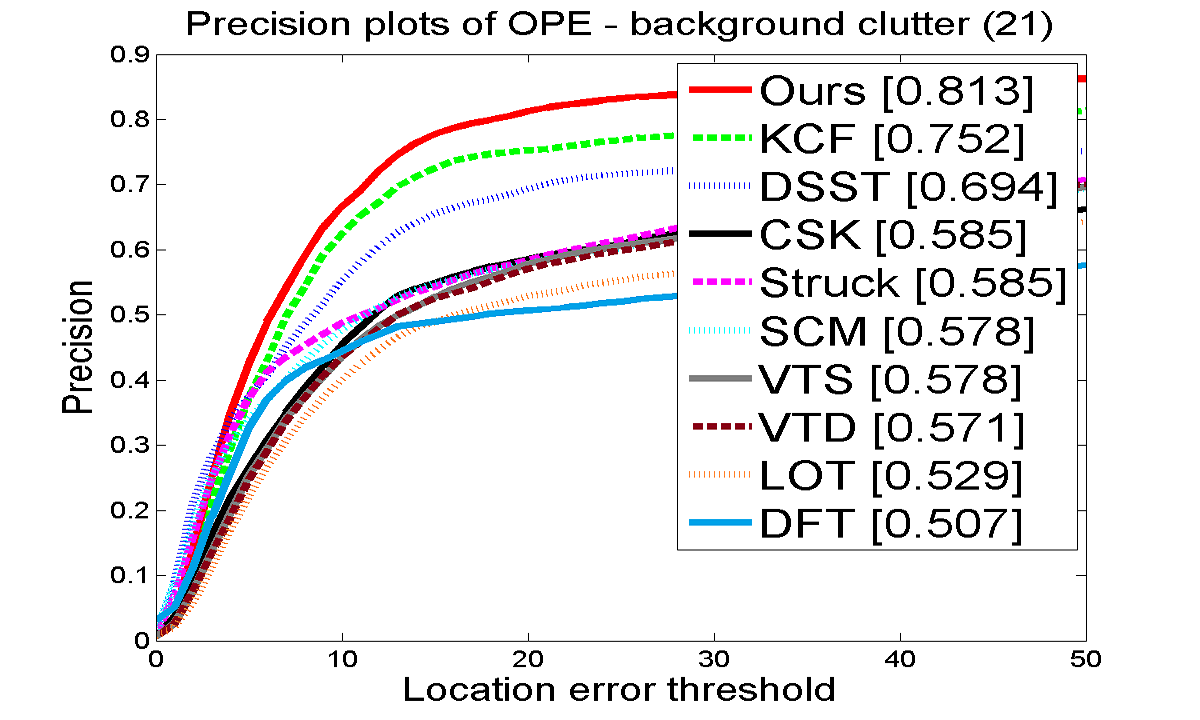}}
\subfigure[Motion blur]{\includegraphics[width=0.245\textwidth]{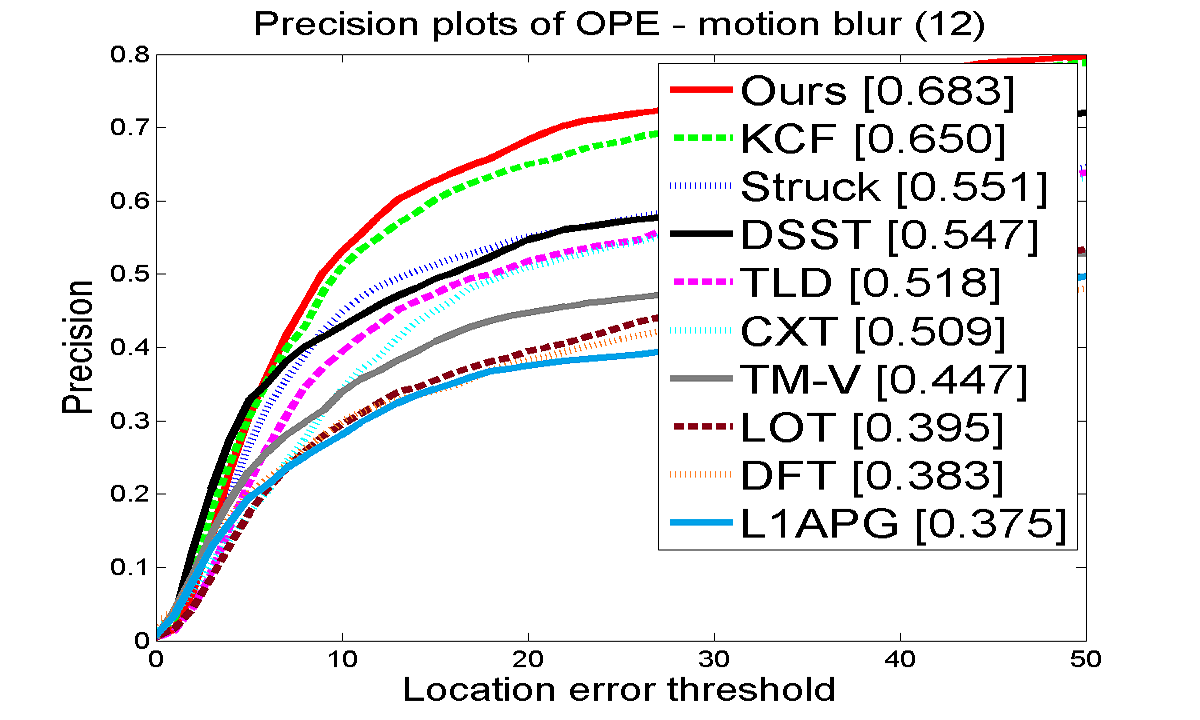}}
\subfigure[Deformation]{\includegraphics[width=0.245\textwidth]{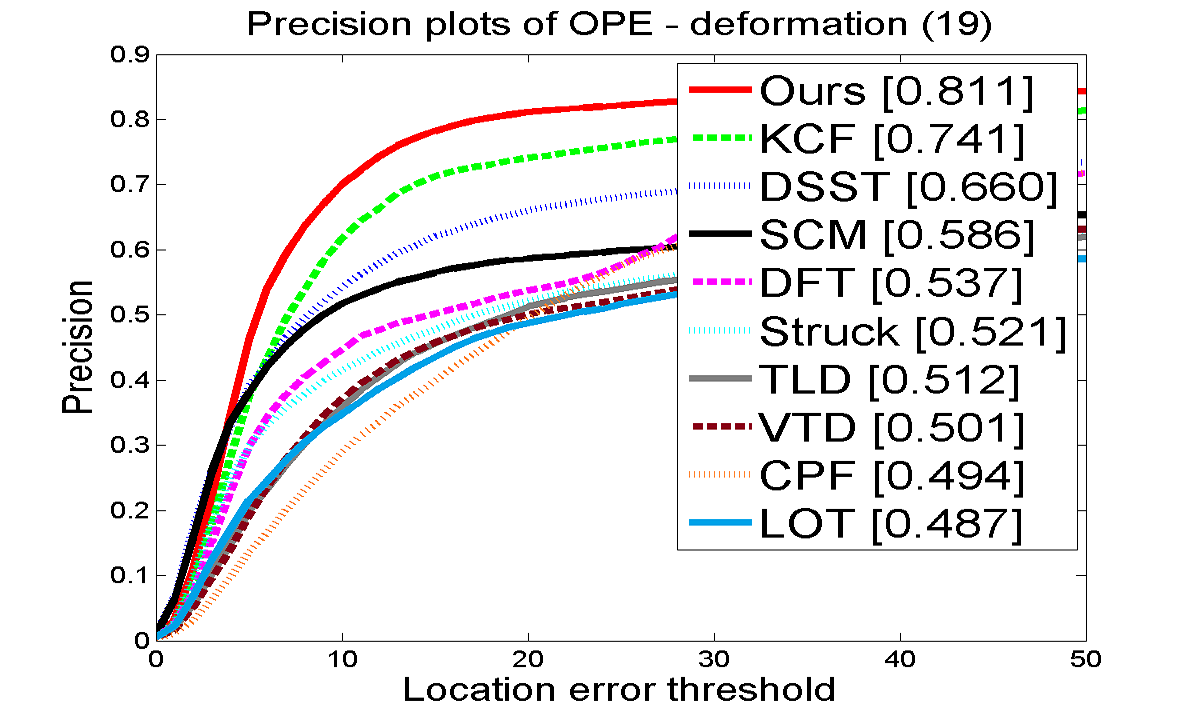}}
\subfigure[Illumination variation]{\includegraphics[width=0.245\textwidth]{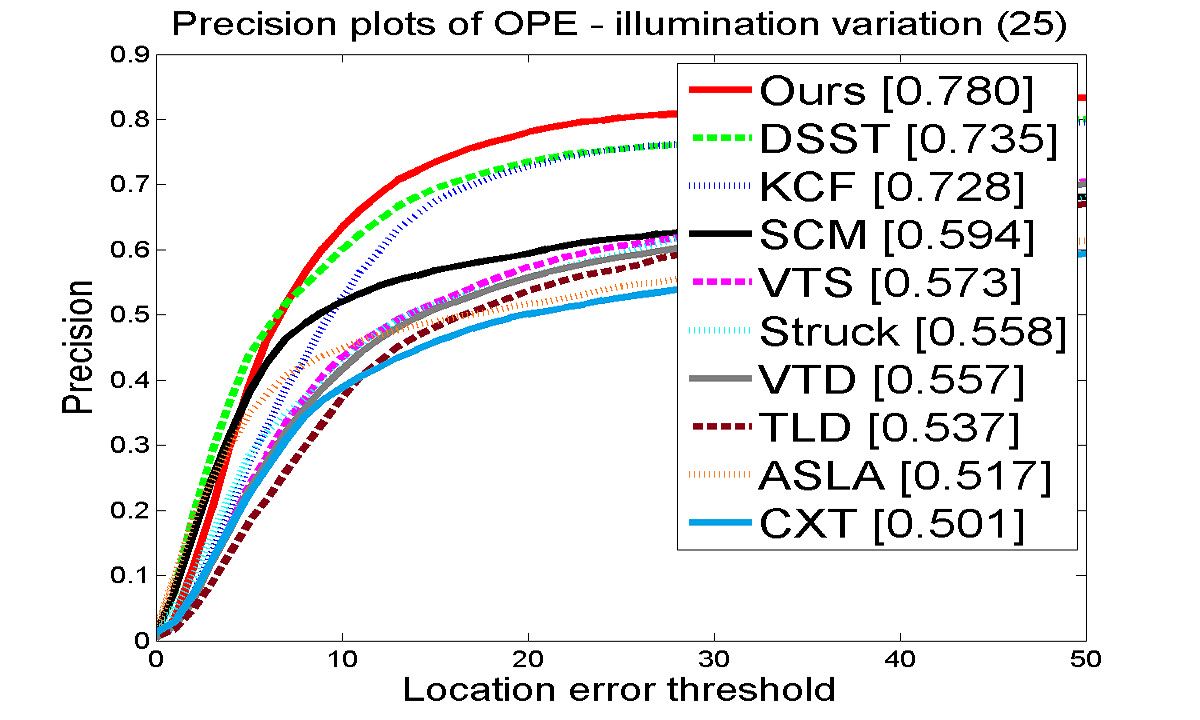}}
\subfigure[In-plane rotation]{\includegraphics[width=0.245\textwidth]{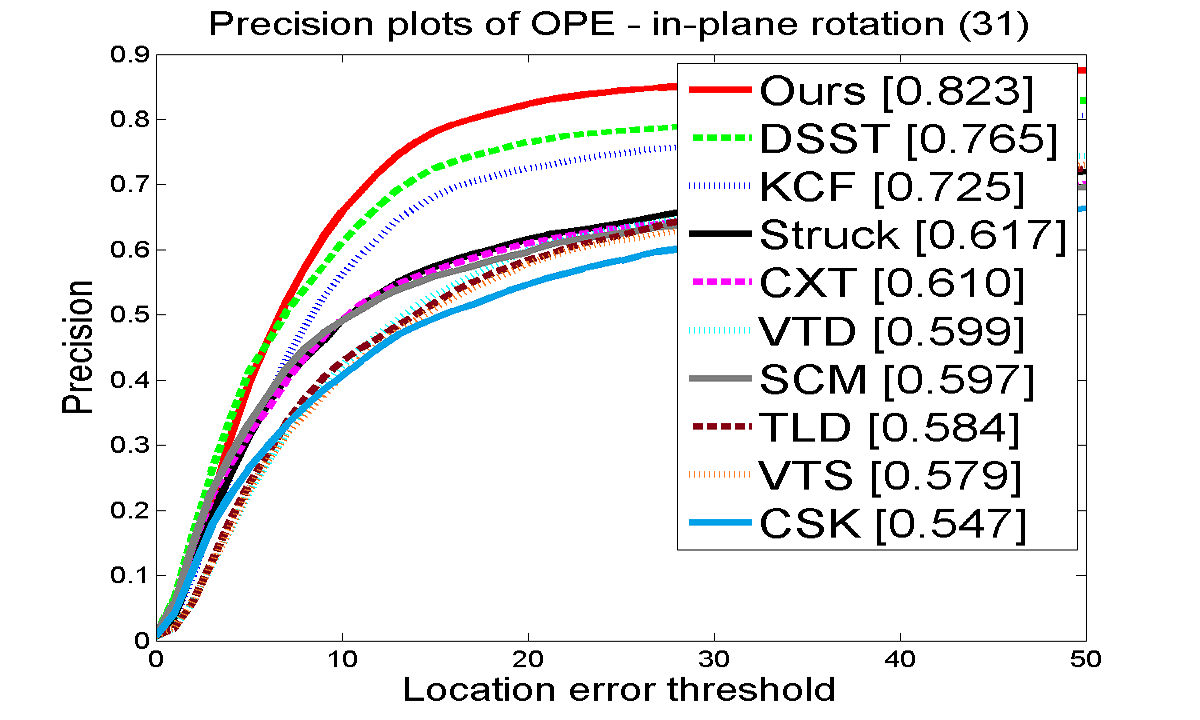}}
\subfigure[Low resolution]{\includegraphics[width=0.245\textwidth]{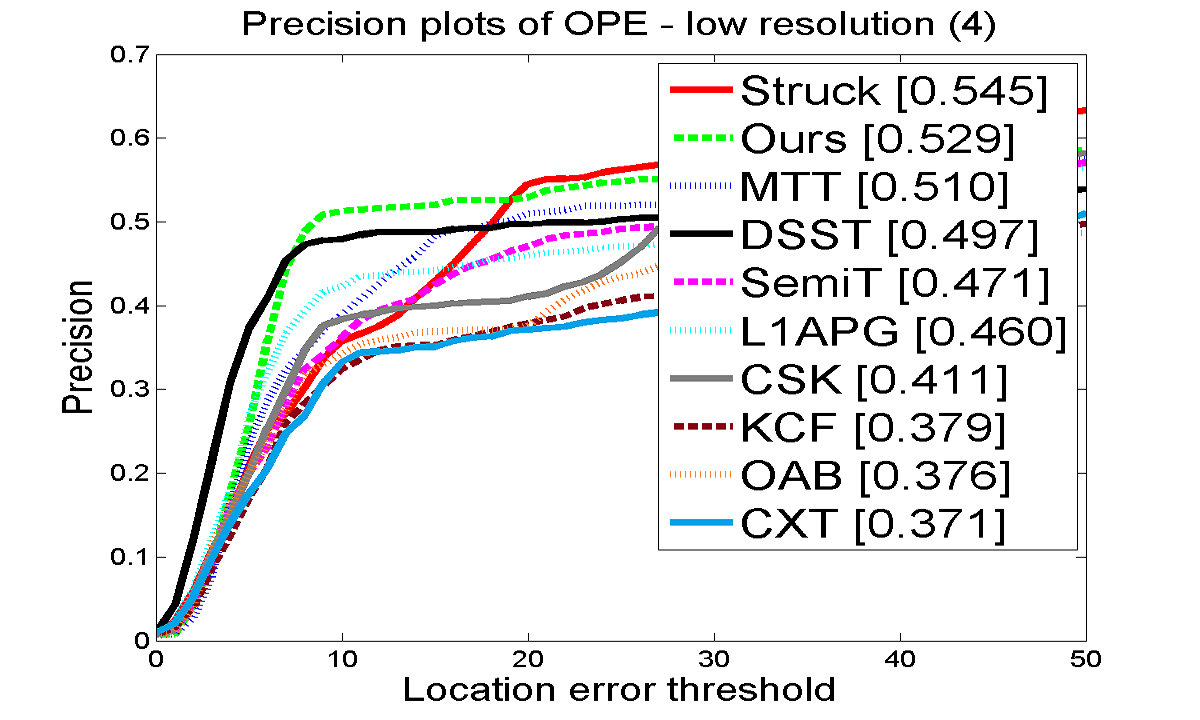}}
\subfigure[Occlusion]{\includegraphics[width=0.245\textwidth]{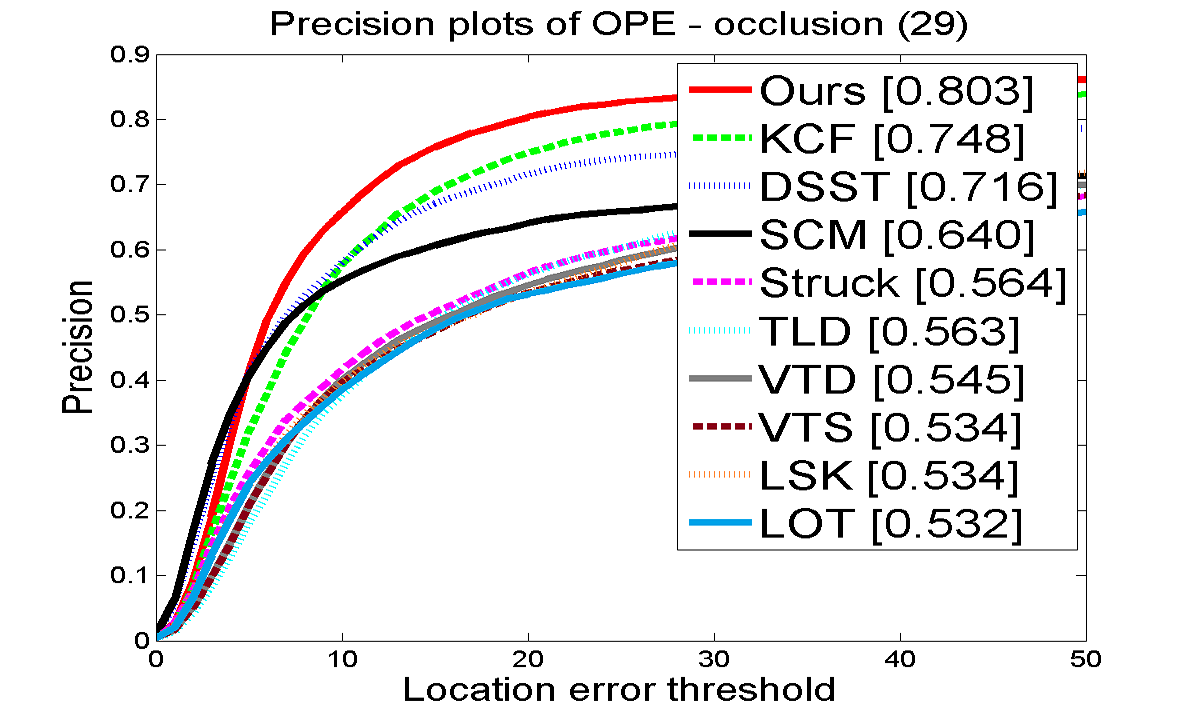}}
\subfigure[Out-of-plane rotation]{\includegraphics[width=0.245\textwidth]{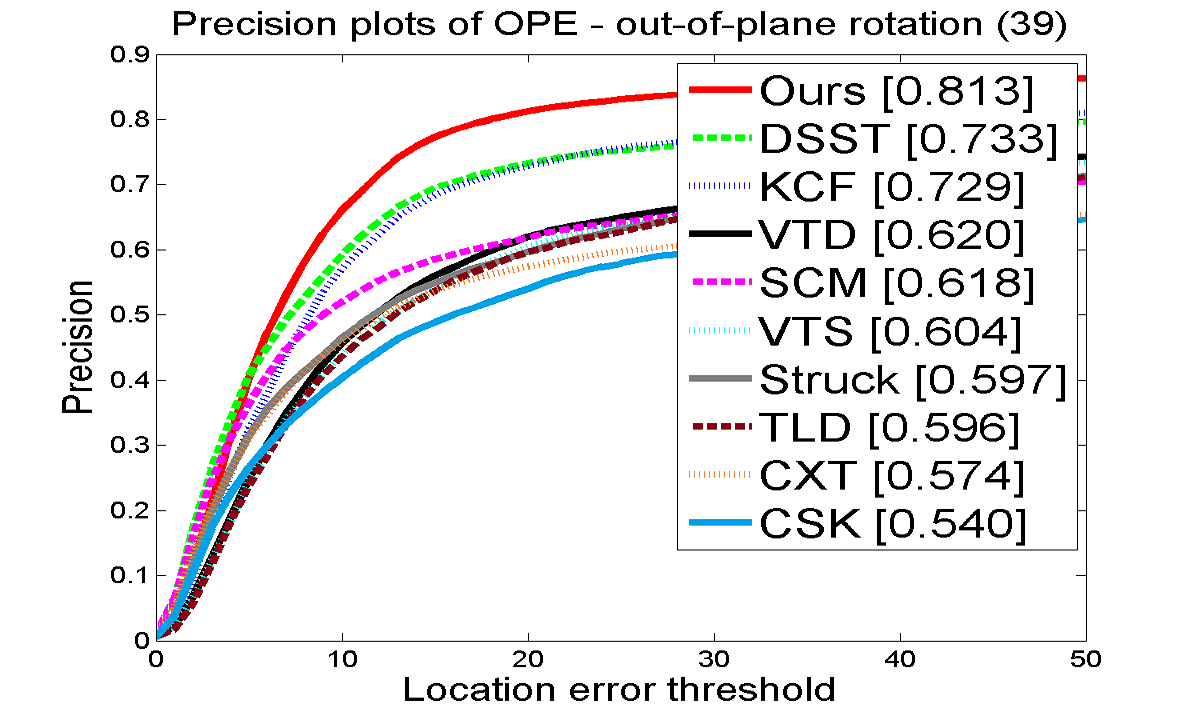}}
\subfigure[Out-of view]{\includegraphics[width=0.245\textwidth]{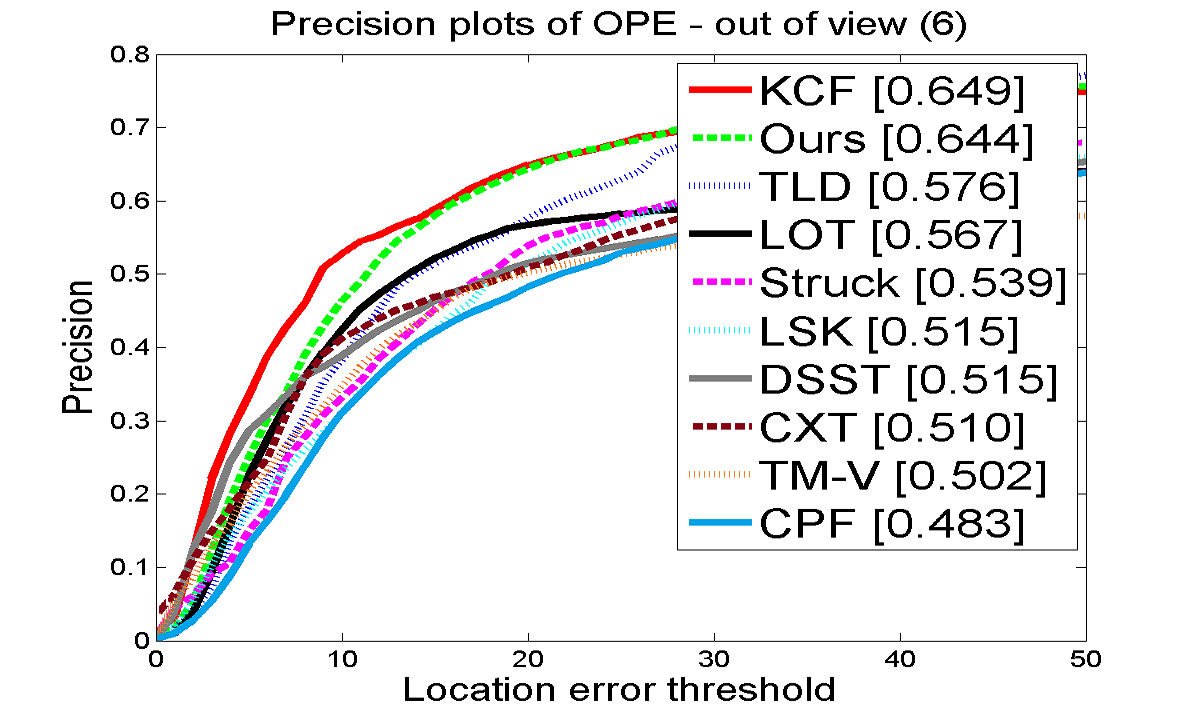}}
\subfigure[Scale variation]{\includegraphics[width=0.245\textwidth]{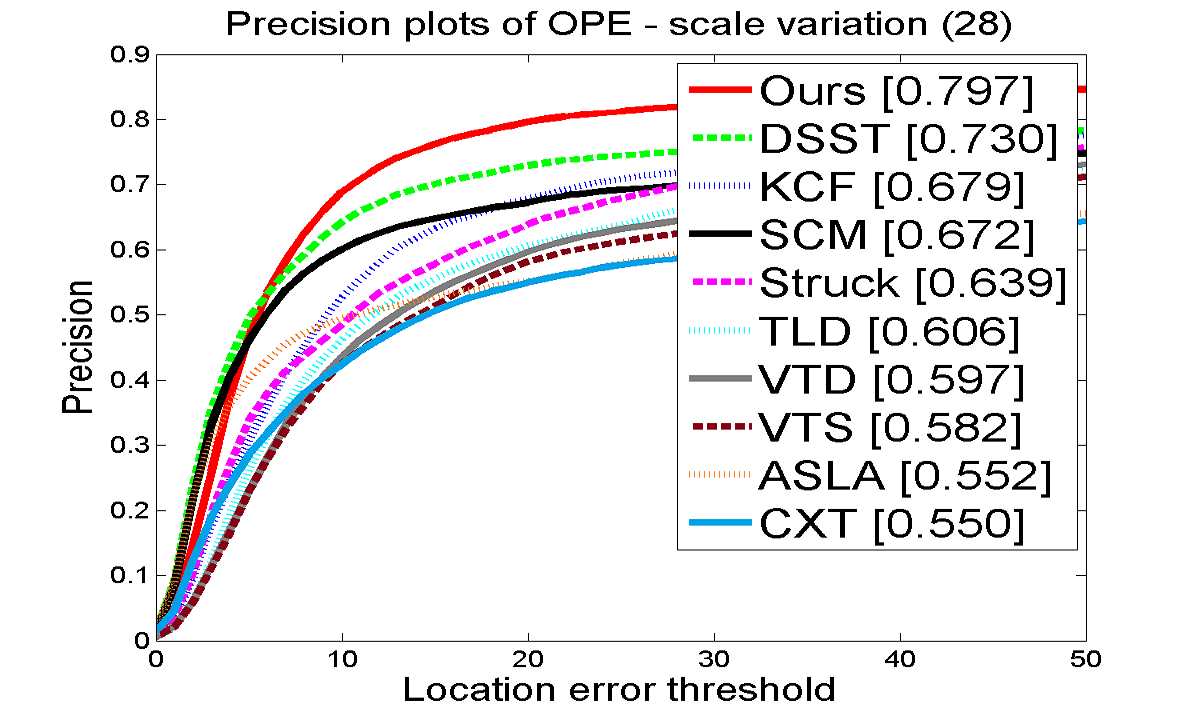}}
\caption{
The precision plots of evaluation of different attributes on OTB2013. The number at the end of the caption of each sub-figure shows how many sequences are included in the corresponding case.
}
\label{fig:opespre}
\end{figure*}

\begin{figure}[!t]
%\centering
\subfigure[Fast motion]{\includegraphics[width=0.245\textwidth]{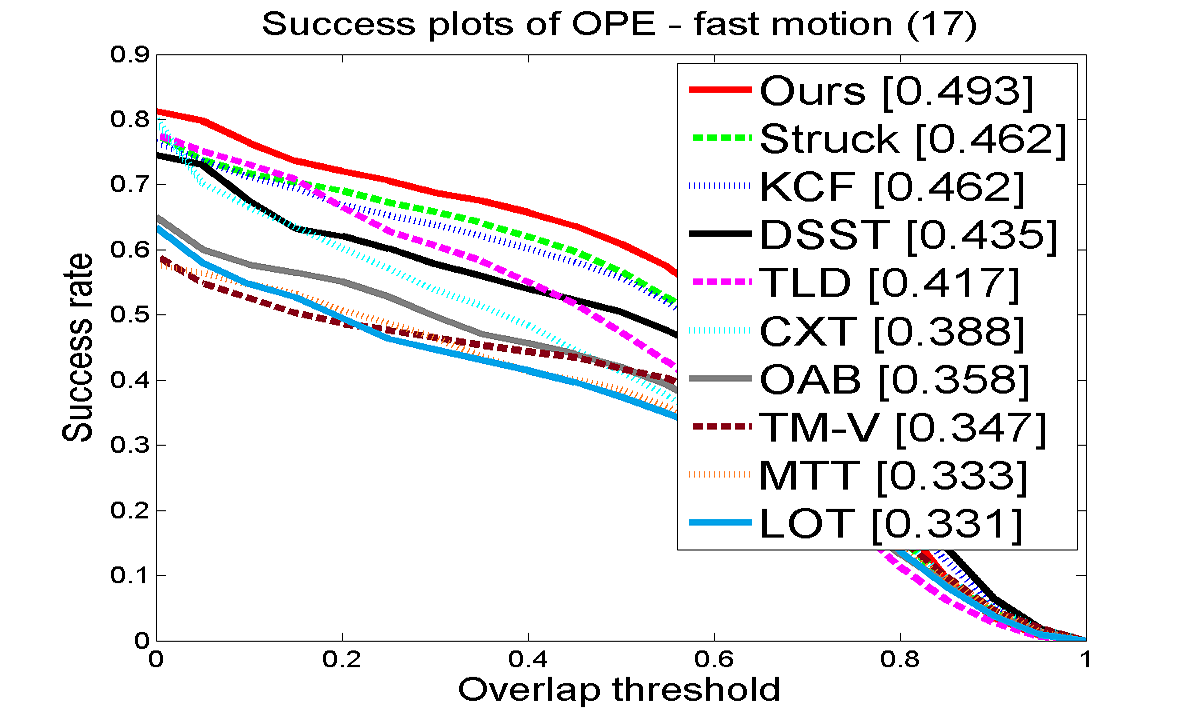}}
\subfigure[Background clutter]{\includegraphics[width=0.245\textwidth]{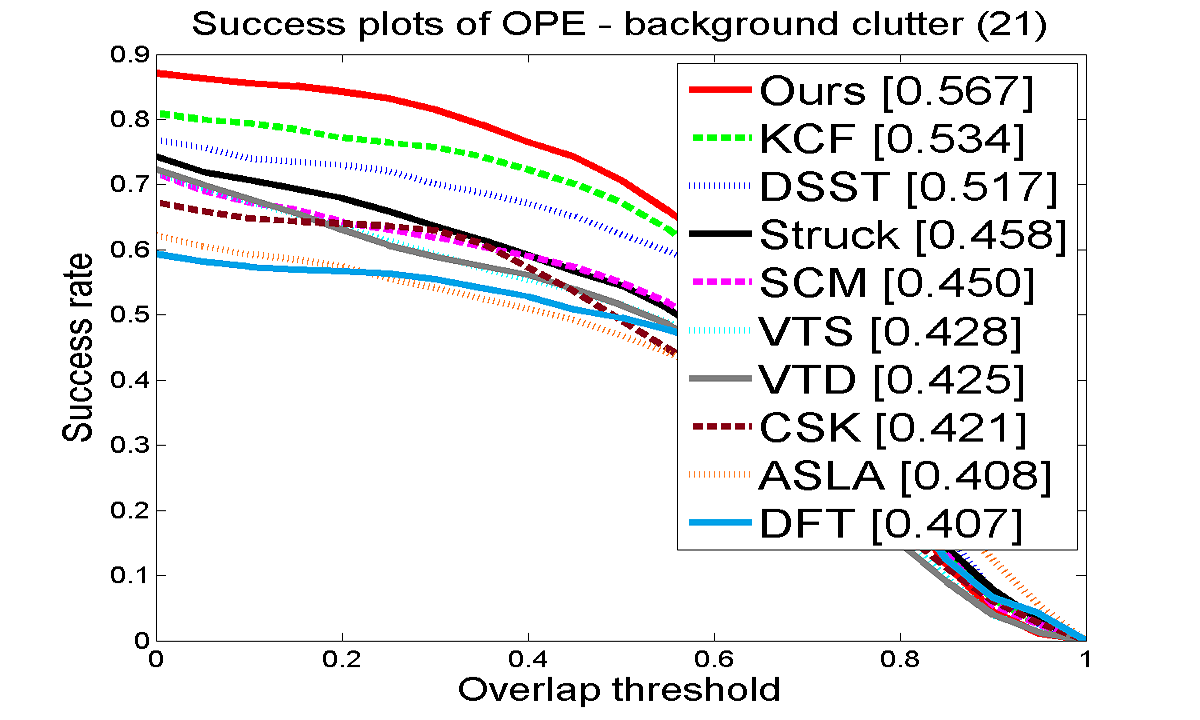}}
\subfigure[Motion blur]{\includegraphics[width=0.245\textwidth]{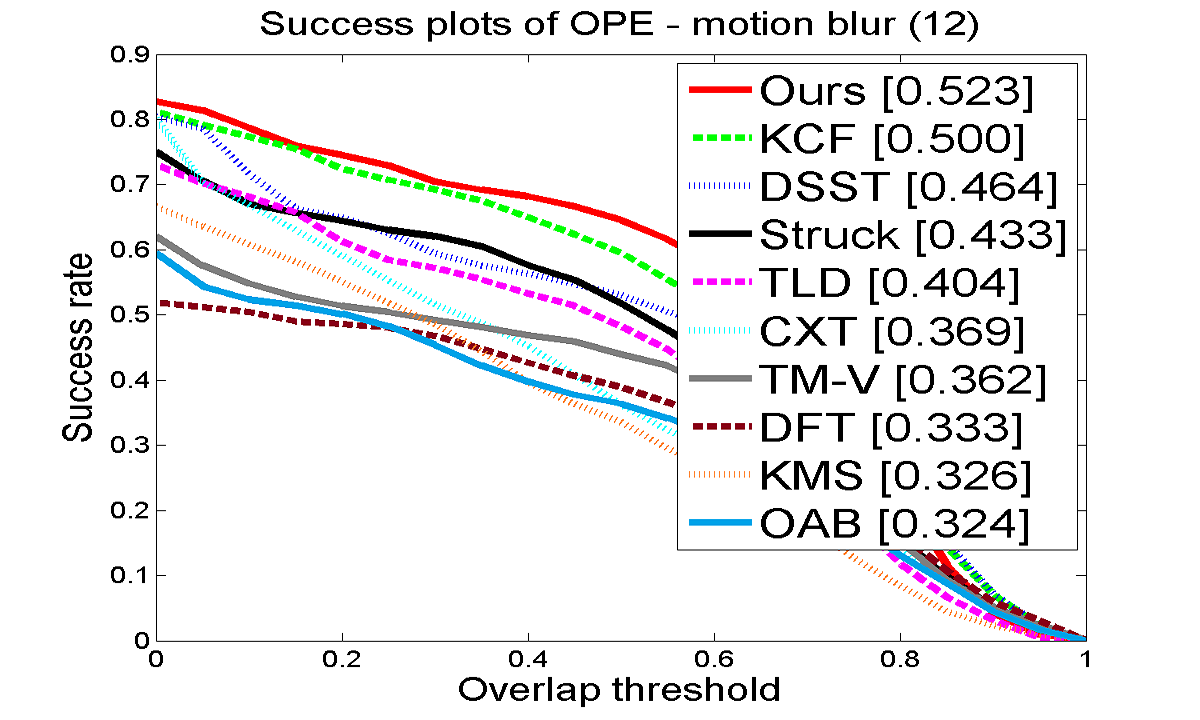}}
\subfigure[Deformation]{\includegraphics[width=0.245\textwidth]{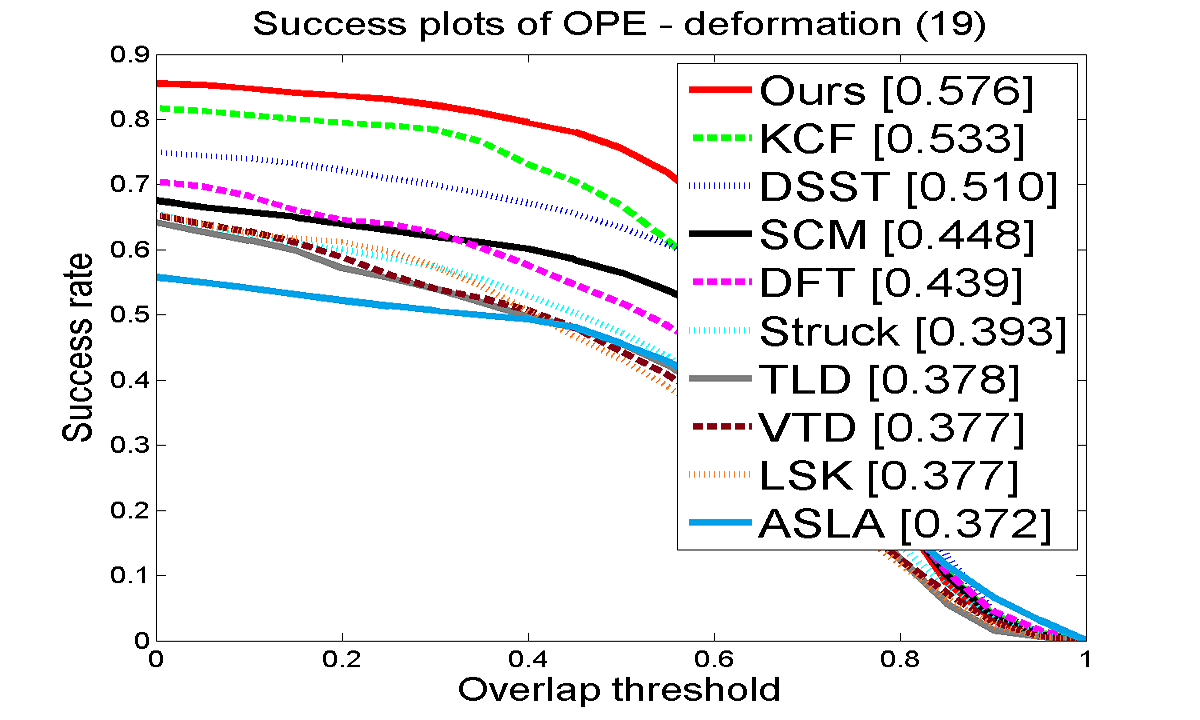}}
\subfigure[Illumination variation]{\includegraphics[width=0.245\textwidth]{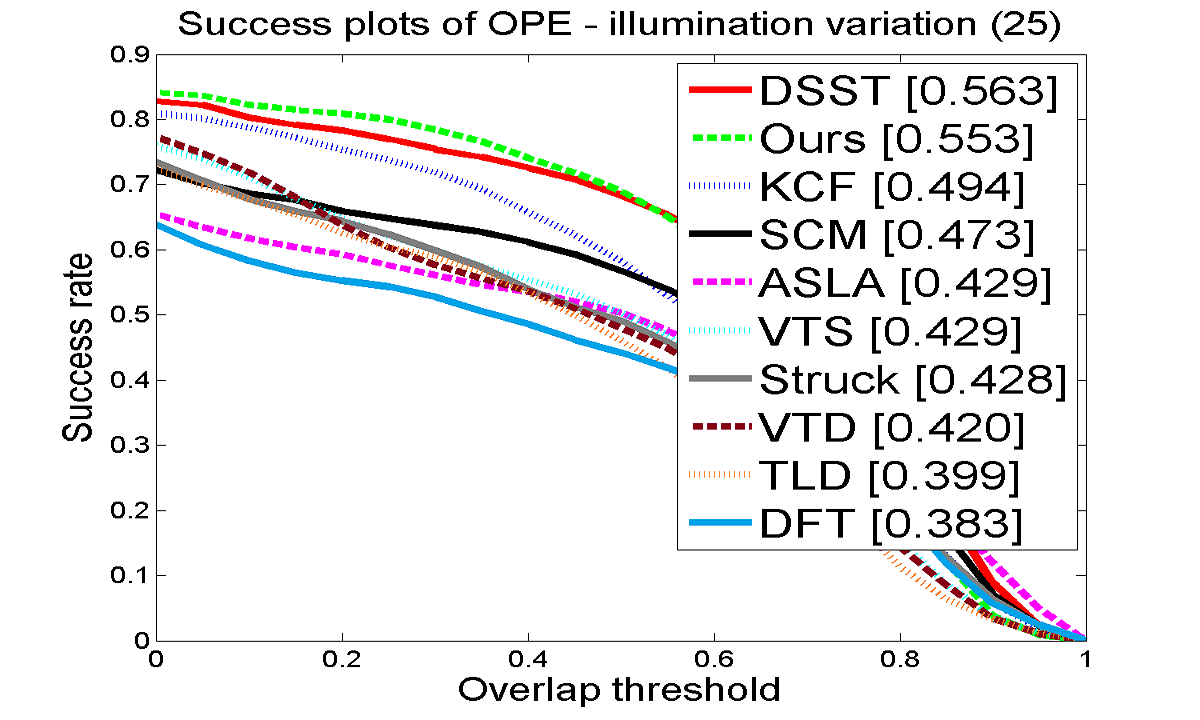}}
\subfigure[In-plane rotation]{\includegraphics[width=0.245\textwidth]{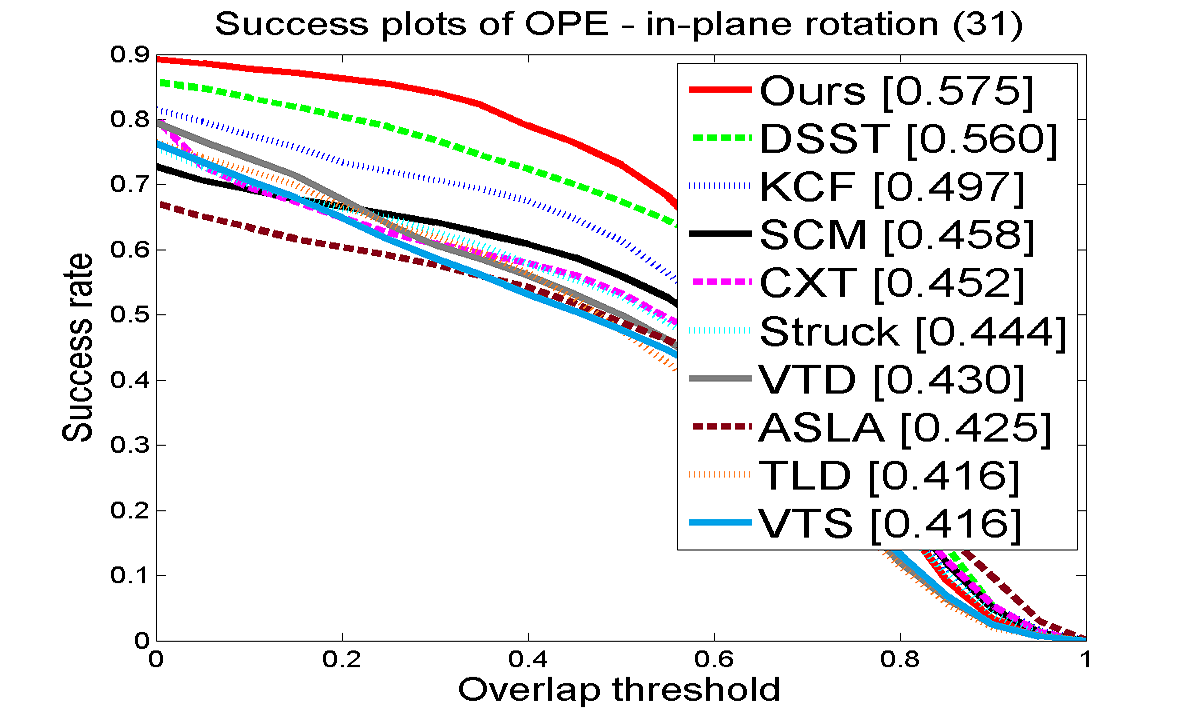}}
\subfigure[Low resolution]{\includegraphics[width=0.245\textwidth]{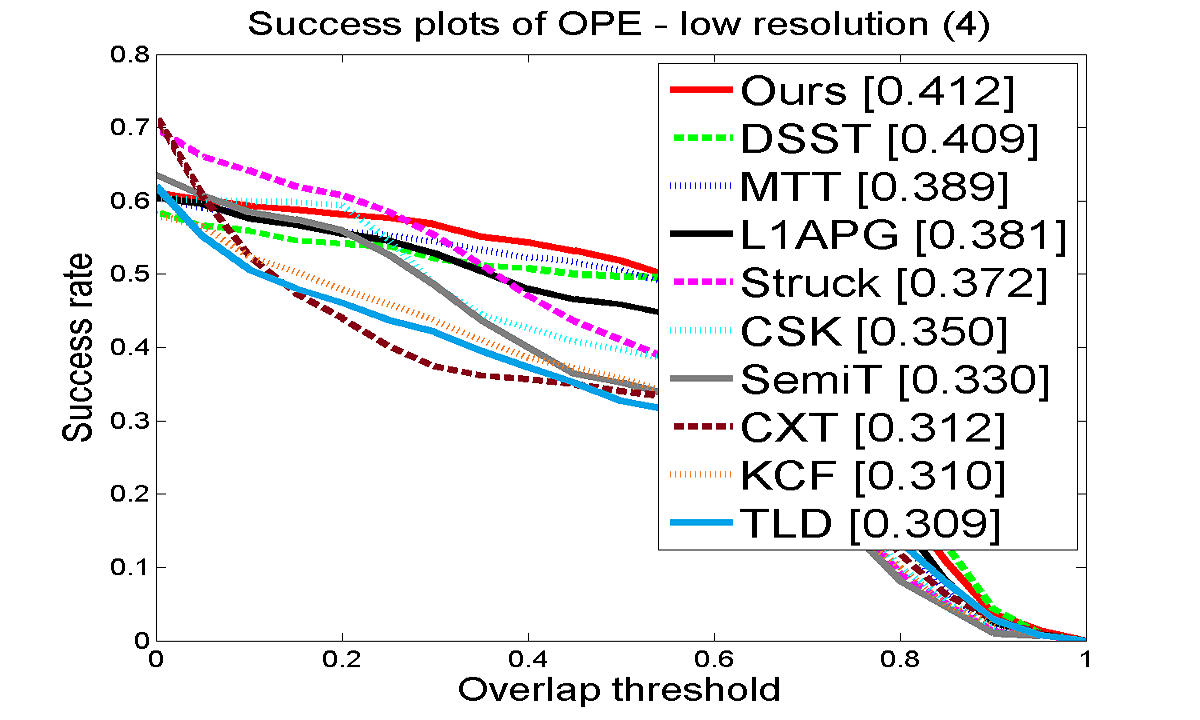}}
\subfigure[Occlusion]{\includegraphics[width=0.245\textwidth]{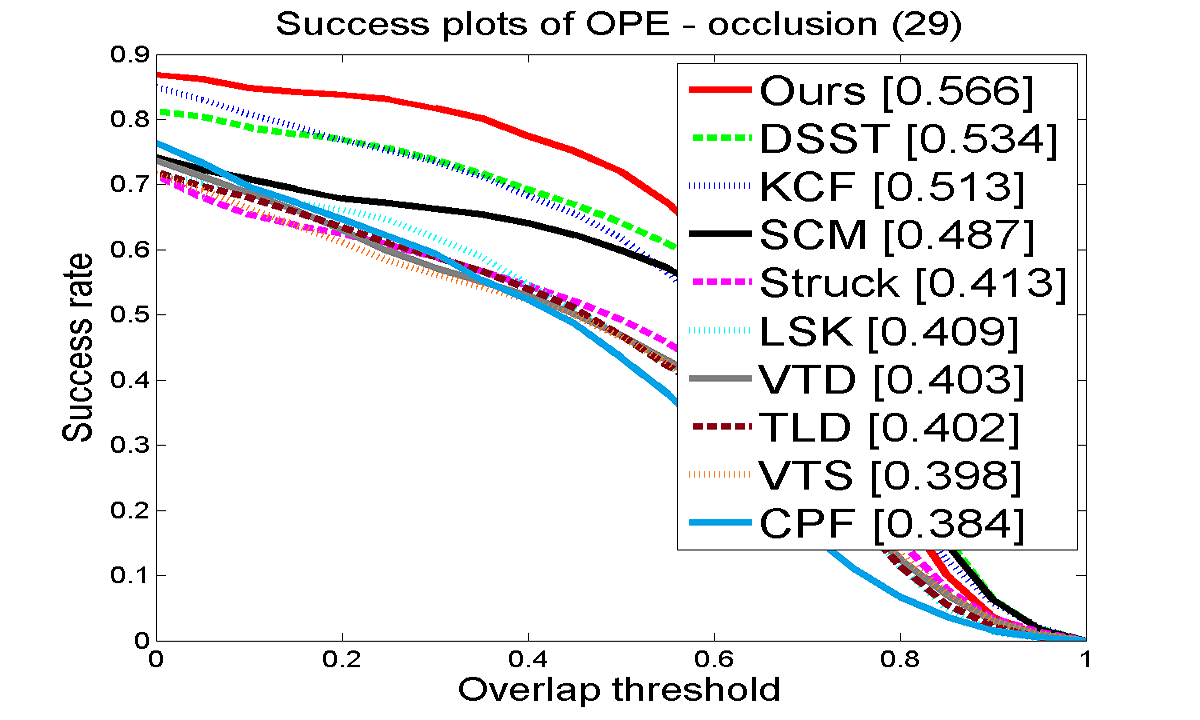}}
\subfigure[Out-of-plane rotation]{\includegraphics[width=0.245\textwidth]{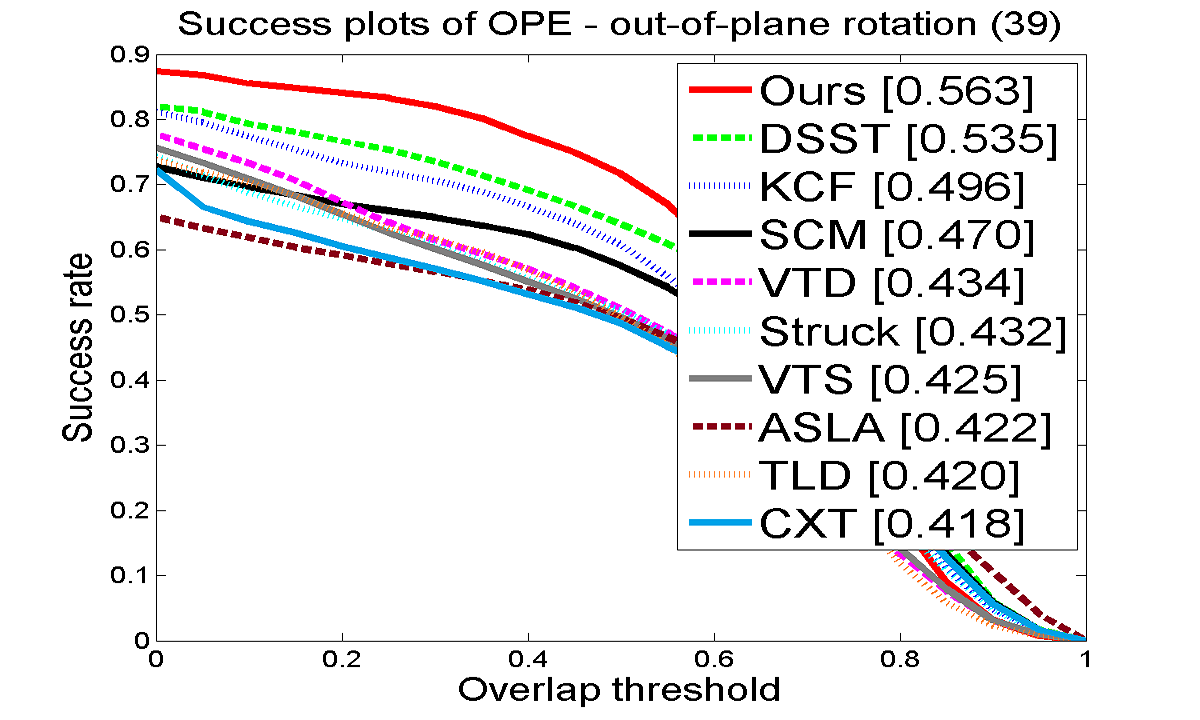}}
\subfigure[Out-of view]{\includegraphics[width=0.245\textwidth]{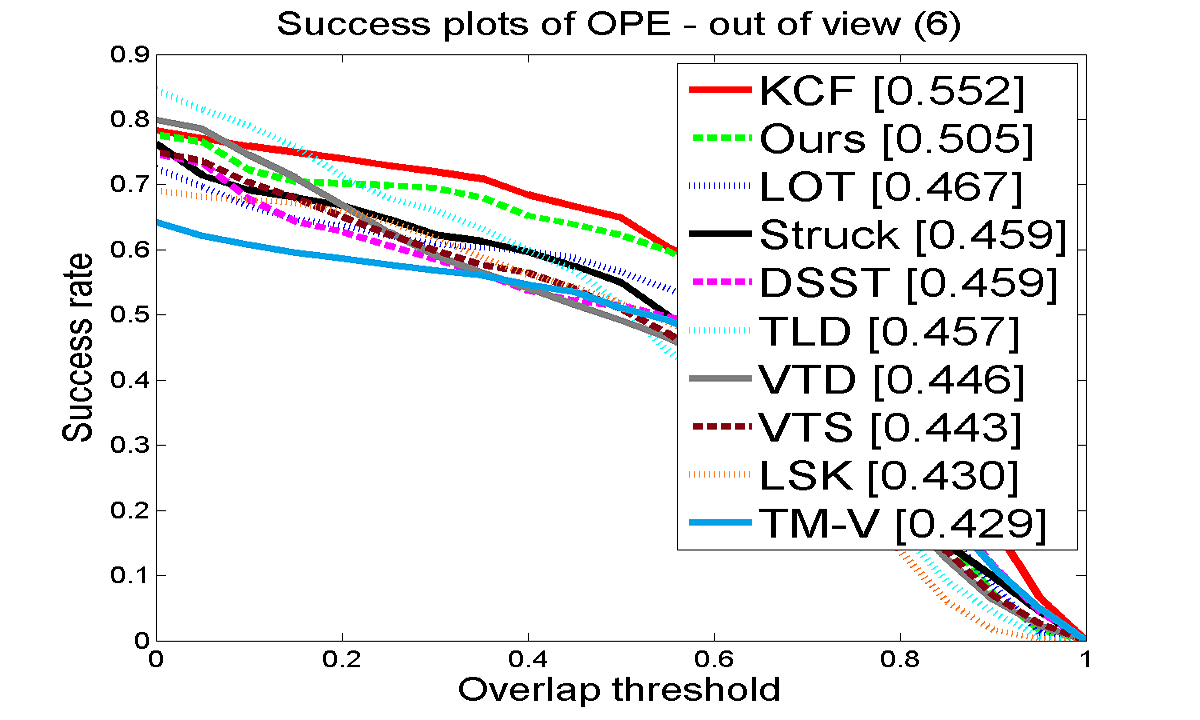}}
\subfigure[Scale variation]{\includegraphics[width=0.245\textwidth]{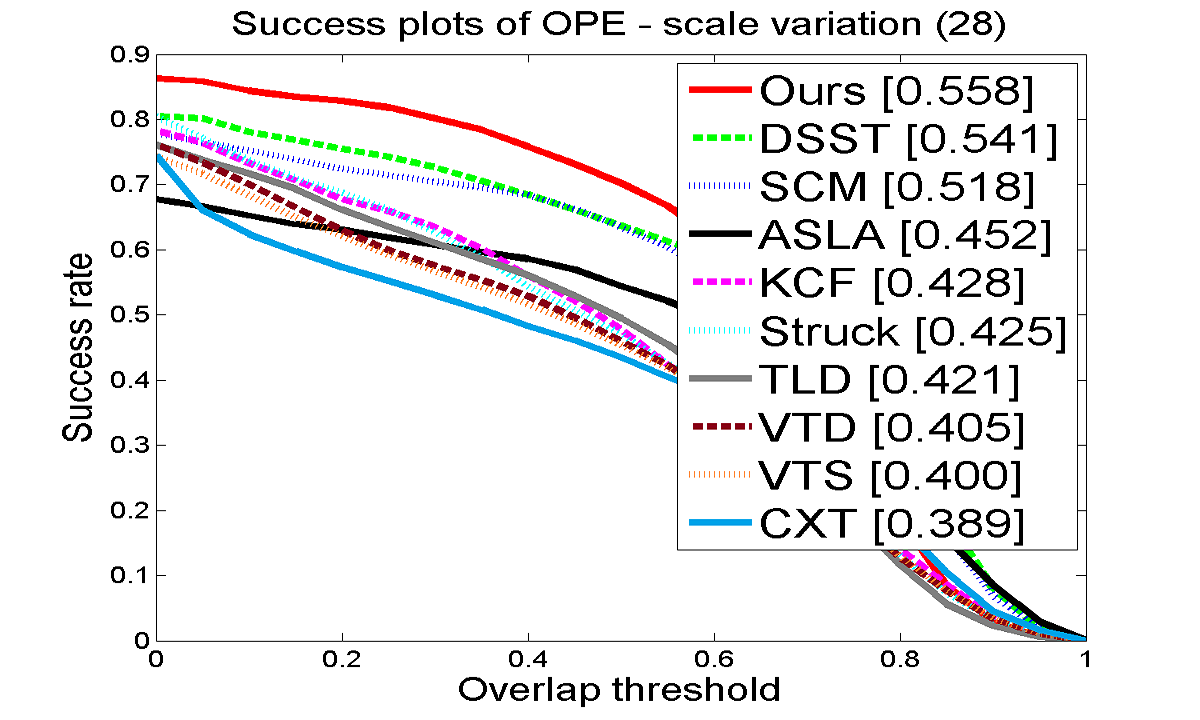}}
\caption{
The success plots of evaluation of different attributes on OTB2013. The number at the end of the caption of each sub-figure shows how many sequences are included in the corresponding case.
}
\label{fig:opessue}
\end{figure}

The OPE performances of the trackers on each attribute are shown in Figures ~\ref{fig:opespre} and \ref{fig:opessue}, which demonstrates the OPE performances of the top ten trackers on the $11$ attributes.
Our proposed tracker achieves the best or the second best performance among all of the trackers compared, and the performance of different attribute groups indicates that our CFPFT tracker is clearly more accurate and robust.
These advantages benefit from the particle filter redetection and the scale evaluation mechanism.

\begin{figure*}[!t]
\centering
\subfigure[The precision plots of SRE]{\includegraphics[width=0.48\textwidth]{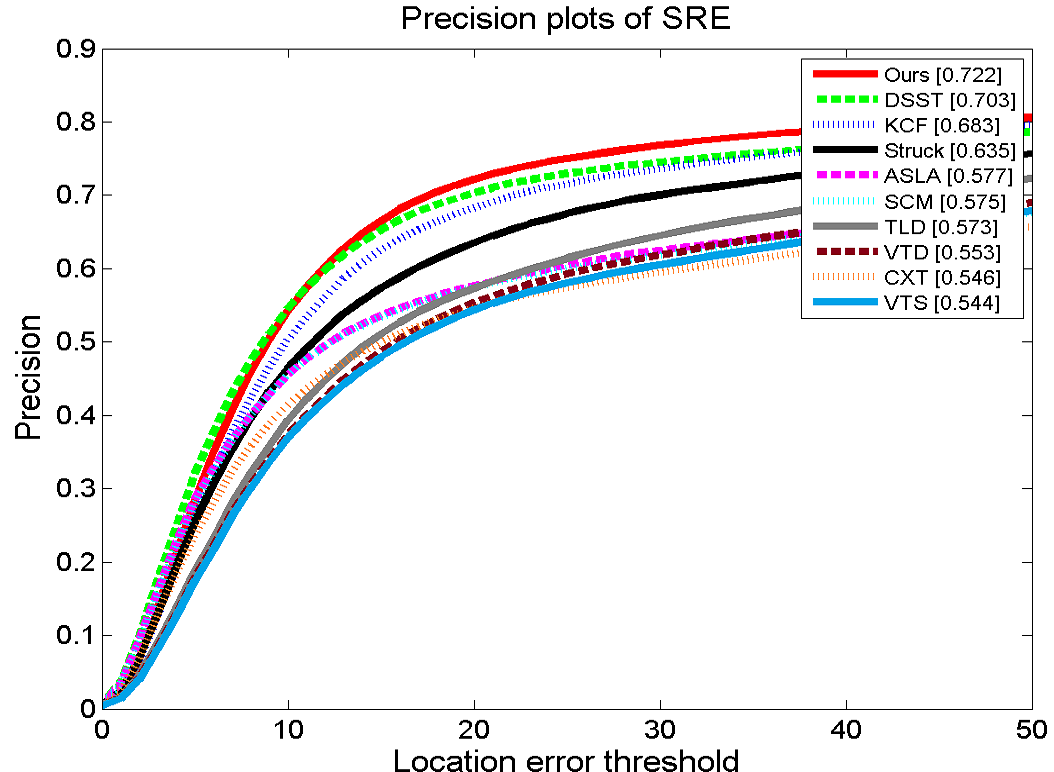}}
\subfigure[The success plots of SRE]{\includegraphics[width=0.48\textwidth]{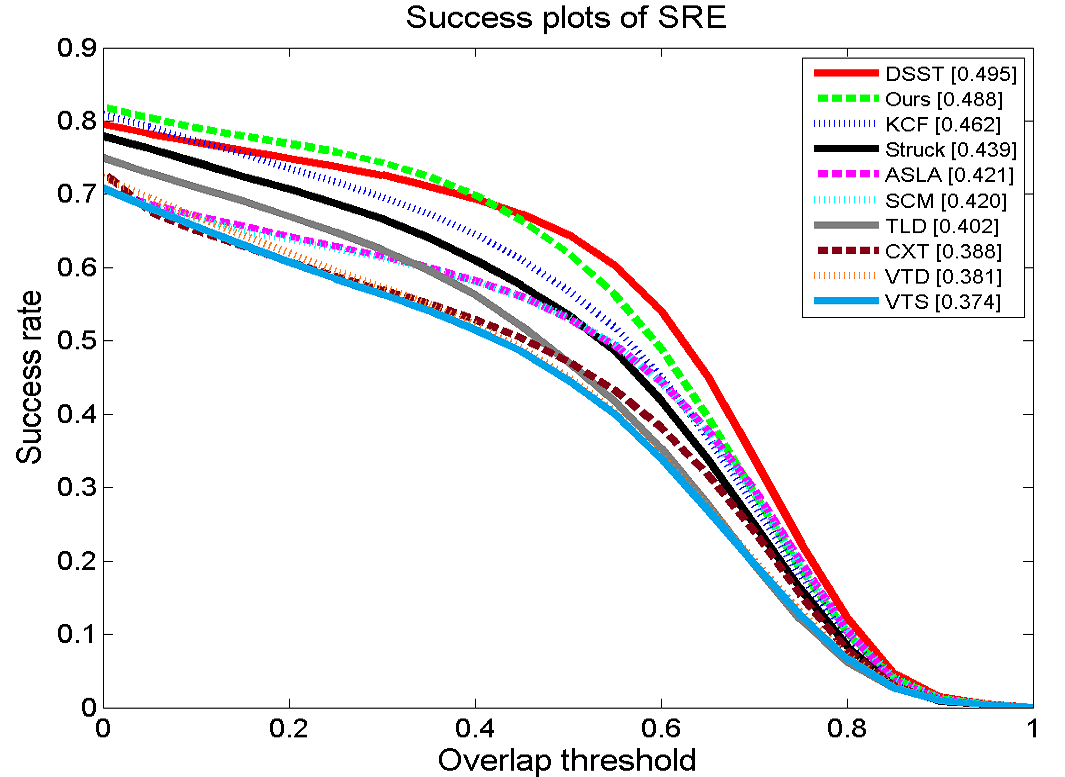}}
\subfigure[The precision plots of TRE]{\includegraphics[width=0.48\textwidth]{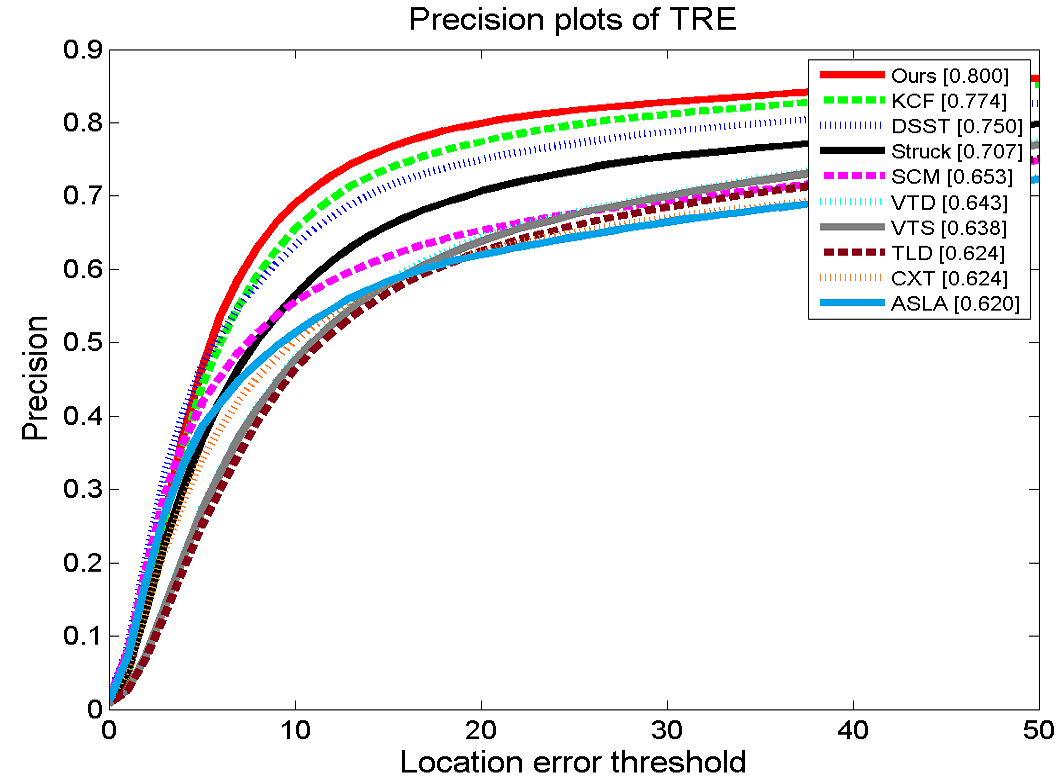}}
\subfigure[The success plots of TRE]{\includegraphics[width=0.48\textwidth]{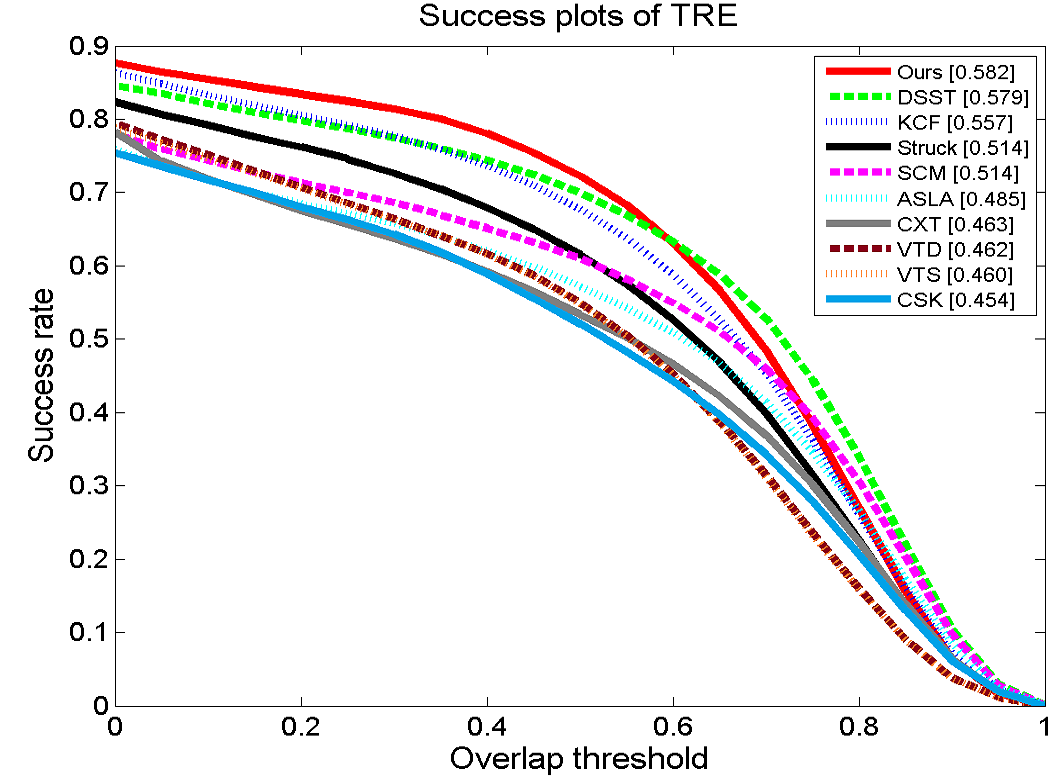}}
\caption{
 The precision and success plots of SRE and TRE on the OTB2013 benchmark.
}
\label{fig:sre}
\end{figure*}

% SRE, TRE evaluation
In order to provide sufficient experimental comparison results to verify the robustness of our CFPFT tracker, we show the overall comparison performance for SRE and TRE in Figure ~\ref{fig:sre}.
The Figures ~\ref{fig:sre}(a-b) shows that our tracker achieves the second best performance on the success plots, close to that of DSST and better than that of KCF. On the precision plots, our CFPFT tracker achieves the best performance and is 2$\%$ higher than that of DSST, which is in the second place.
From Figures ~\ref{fig:sre}(c-d), we see both the precision and the success plots shows that our tracker achieves the best performance. The results for TRE show the robustness of our tracker to the initialization in the first frame by shifting or scaling the ground truth.
Because our CFPFT is based on KCF, the results show the robustness of the redetection mechanism.
To summarize briefly, the CFPFT tracker is effective and achieves promising results on the visual tracking benchmark OTB2013.

\subsubsection{Evaluation with OTB2015}
To further evaluate the performance of our proposed approach, in this section, we compare the performance of our CFPFT tracker with some state-of-the-art trackers, including
TGPR\cite{Gao2014Transfer}, DSST\cite{Springer:DSST}, KCF\cite{IEEE:KCF}, SCM\cite{Yang2012Robust}, Struck\cite{Hare2012Struck}, CNN-SVM\cite{CNN-SVM}, CNT\cite{CNT}, CFNet-conv1\cite{CFNet}, and HDT\cite{HDT}, and the last four trackers are based on deep learning theory.
Unlike other methods, deep neural network based trackers extract the features by utilizing deep learning.

\begin{figure*}[!t]
\centering
\subfigure[The precision plots of OPE]{\includegraphics[width=0.48\textwidth]{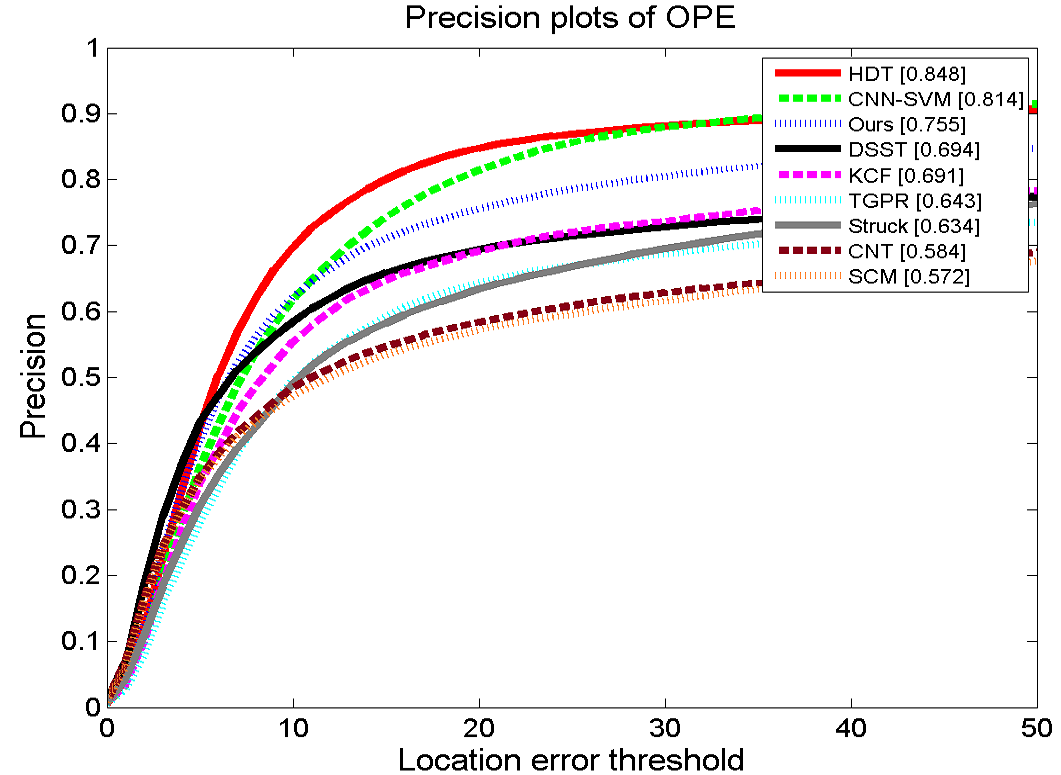}}
\subfigure[The success plots of OPE]{\includegraphics[width=0.48\textwidth]{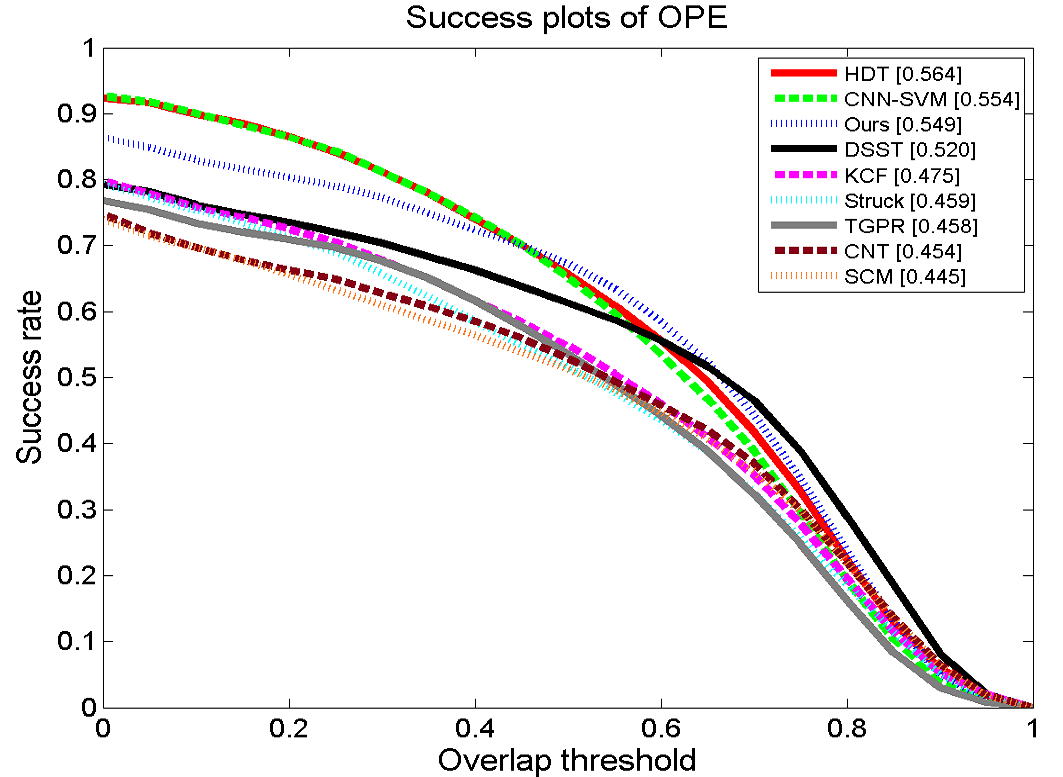}}
\caption{
The precision and success plots of OPE on OTB2015 over 100 standard benchmark sequences.
}
\label{fig:otb100}
\end{figure*}

Figure ~\ref{fig:otb100} shows the precision and the success plots of our CFPFT tracker and the eight state-of-the-art trackers on the OTB2015 dataset.
From that, we can see the success rate and the precision rate of our approach are just below those of HDT\cite{HDT} and CNN-SVM\cite{CNN-SVM}, which uses the deep feature. Meanwhile, it is obvious that our CFPFT tracker achieves promising performance and outperforms other six trackers which include a deep learning based tracker\cite{CNT}, two correlation filter based trackers\cite{IEEE:KCF,Springer:DSST} and three representative trackers\cite{Gao2014Transfer,Yang2012Robust,Hare2012Struck}.

We further analyze the performance of CFPFT for different attributes in OTB2015\cite{Wu2015Object}.
Table ~\ref{table:otb15} shows the comparisons of CFPFT with  eight other state-of-the-art tracking algorithms on these $11$ attributes.
In terms of distance precision rates (DPR), CFPFT achieves the best or close to the best results for all $11$ attributes.
Compared with the deep learning based tracker\cite{CNT}, CFPFT can locate the object  better in videos but performs worse than the other two deep learning based trackers\cite{CNN-SVM,HDT}.
Compared with other trackers, including correlation filter based trackers and traditional trackers, CFPFT can locate the object better.
On the other hand, CFPFT also achieves the best or close to the best results of overlap success rates (OSR) for all $11$  attributes.
Compared with all eight state-of-the-art trackers, CFPFT performs more robustly of fast motion, background cluttered, low resolution, occlusion and scale variation.
While compared with correlation filters based trackers\cite{Springer:DSST,IEEE:KCF} and the traditional method\cite{Gao2014Transfer,Yang2012Robust}, CFPFT performs more robustly for all $11$  attributes with the help of the redetection mechanism and the scale evaluation mechanism.

\begin{table*}[!t]
\tiny
%\scriptsize
\centering
    \caption{ Average precision and success scores of our CFPFT and other trackers on OTB2015 for different attributes.
    The value format of each table cell is " DPR/OSR ($\%$)
    for 11 attributes".}
    \begin{tabular}{c|ccccccccc}  %{|c|c|c|c|c|c|c|c|c|}
\hline
Attribute  & CFPFT & KCF & DSST & HDT & CNN-SVM & CNT & TGPR & SCM & Struck\\
     %\midrule
     \hline
BC  & $75.4$/$54.1$ & $71.2$/$49.7$ & $70.4$/$52.1$ & $84.4$/$57.8$ & $77.6$/$54.8$ & $62.4$/$49.0$ & $59.3$/$42.8$ & $57.7$/$46.2$ & $54.7$/$42.6$\\
DEF  & $68.9$/$49.4$ & $61.7$/$43.6$ & $57.0$/$43.4$ & $82.1$/$53.4$ & $79.3$/$54.7$ & $52.4$/$39.8$ & $63.0$/$45.5$ & $52.4$/$40.2$ & $52.7$/$38.3$\\
FM  & $67.7$/$52.1$ & $61.9$/$44.9$ & $58.3$/$47.0$ & $80.2$/$55.0$ & $74.2$/$52.9$ & $37.7$/$32.6$ & $50.7$/$39.8$ & $34.9$/$31.9$ & $60.0$/$45.0$\\
IPR  & $75.9$/$53.0$ & $69.3$/$46.5$ & $71.3$/$51.0$ & $84.4$/$55.5$ & $81.3$/$51.8$ & $55.3$/$41.3$ & $65.9$/$46.2$ & $54.3$/$40.8$ & $62.5$/$44.6$\\
IV  & $76.7$/$56.7$ & $70.7$/$47.4$ & $72.6$/$55.9$ & $82.0$/$53.5$ & $79.5$/$53.7$ & $56.7$/$46.2$ & $63.3$/$45.2$ & $59.7$/$48.7$ & $54.9$/$42.0$\\
LR  & $65.6$/$46.6$ & $54.5$/$30.6$ & $58.1$/$38.9$ & $76.6$/$42.0$ & $79.0$/$41.9$ & $57.9$/$41.0$ & $62.9$/$37.8$ & $55.8$/$38.1$ & $62.8$/$34.7$\\
MB  & $70.7$/$55.7$ & $61.7$/$45.7$ & $59.7$/$49.1$ & $79.4$/$56.3$ & $76.7$/$56.8$ & $36.9$/$35.8$ & $50.8$/$40.9$ & $31.6$/$30.8$ & $58.0$/$45.1$\\
OCC  & $69.2$/$50.1$ & $62.1$/$43.8$ & $60.9$/$46.0$ & $77.4$/$52.8$ & $73.0$/$51.5$ & $55.4$/$43.4$ & $59.4$/$42.9$ & $54.9$/$42.2$ & $52.4$/$38.7$\\
OPR  & $74.4$/$52.7$ & $67.0$/$45.0$ & $66.5$/$48.1$ & $80.5$/$53.3$ & $79.8$/$54.8$ & $57.6$/$43.6$ & $64.2$/$45.5$ & $56.9$/$43.1$ & $59.3$/$42.4$\\
OV  & $56.2$/$41.1$ & $49.8$/$39.4$ & $48.0$/$38.5$ & $66.3$/$47.2$ & $65.0$/$48.8$ & $37.4$/$34.1$ & $49.3$/$37.3$ & $42.3$/$33.3$ & $46.0$/$35.7$\\
SV  & $71.2$/$51.2$ & $63.9$/$39.9$ & $66.4$/$48.5$ & $81.1$/$48.9$ & $79.0$/$49.3$ & $53.1$/$41.7$ & $59.1$/$39.9$ & $56.5$/$43.6$ & $59.7$/$40.3$\\
\hline
average  & $75.5$/$54.9$ & $69.1$/$47.5$ & $69.4$/$52.0$ & $84.8$/$56.4$ & $81.4$/$55.4$ & $58.4$/$45.4$ & $64.3$/$45.8$ & $57.2$/$44.5$ & $63.4$/$45.9$\\
     \hline

     \end{tabular}\label{table:otb15}
\end{table*}

In addition, we use OTB-2013/50/2015 datasets to perform a quantitative comparison of DPR at $20$ pixels and OSR at $0.5$ pixel in Table.~\ref{table:otb}. It shows that our CFPFT outperforms other state-of-the-art trackers at both rates.
On OTB2013 dataset, our tracker achieves a DPR of $82.1\%$ and an OSR of $58.4\%$ and performs slightly worse than HDT\cite{HDT}, which is a deep learning based tracker.
Compared to CFNet-conv1\cite{CFNet}, which is deep learning method, our approach also achieves better DPR and OSR results.
On OTB50 dataset, our tracker achieves a DPR of $69.8\%$ and an OSR of $49.2\%$, achieves the best performance except for the deep learning based tracker HDT\cite{HDT}.
Though CNT\cite{CNT} and CFNet-conv1\cite{CFNet} utilize deep features to represent the appearance of an object, our approach performs better than both of them in term of DPR and OSR results.
Meanwhile, on OTB2015 dataset, our tracker achieves a DPR of $75.5\%$ and an OSR of $54.9\%$, showing better performance than the correlation filters based trackers\cite{Springer:DSST,IEEE:KCF,STC}, the representative trackers\cite{Gao2014Transfer,Yang2012Robust,Hare2012Struck} and the deep learning based trackers\cite{CNT,CFNet}. All of these good results are benefited from the particle resampling mechanism and the simple scale evaluation strategy.
From all the experiment results, we can see that our tracker achieves a good and promising performance.

\begin{table*}[!t]
\tiny
%\scriptsize
\centering
    \caption{Comparisons with state-of-the-art tracking methods on OTB-2013/50/2015\cite{IEEE:OTB2013,Wu2015Object}. Our CFPFT outperforms the existing approaches in term of DPR and OSR ($\%$).}
    \begin{tabular}{c|c|c|ccc|ccc|ccc}  %{|c|c|c|c|c|c|c|c|c|}
\hline
     \multirow{2}{*}{Dataset } %&  \multirow{2}{*}{Evaluation Criterion}
    & Evaluation &  CFPFT &\multicolumn{3}{c|}{Correlation filters trackers} &\multicolumn{3}{c|}{Deep learning trackers} &\multicolumn{3}{c}{Representative trackers}\\
   \cline{4-12}
    & Criterion  &  (Ours) & STC & KCF & DSST & CFNet-conv1 & HDT  & CNT & TGPR & SCM & Struck\\
     \hline
    \multirow{2}{*}{OTB-2013}
    & DPR & $82.1$ & $54.7$ & $74.0$ & $73.7$ & $77.6$ & $88.9$ &  $72.3$ & $76.6$ & $64.9$ & $65.6$\\
    & OSR & $58.4$ & $34.7$ & $51.4$ & $55.4$ & $57.8$ & $60.3$  & $54.5$ & $52.9$ & $49.9$ & $47.4$\\
    \hline
   \multirow{2}{*}{OTB-50}
    & DPR & $69.8$ & $43.1$ & $61.1$ & $62.7$ & $65.3$ & $80.4$ & $50.1$ & $61.2$ & $48.1$ & $52.9$\\
    & OSR & $49.2$ & $27.4$ & $40.3$ & $46.4$ & $48.8$ & $51.5$ & $36.9$ & $42.9$ & $36.4$ & $37.6$\\
    \hline
    \multirow{2}{*}{OTB-2015}
    & DPR & $75.5$ & $50.7$ & $69.1$ & $69.4$ & $71.3$ & $84.8$ & $58.4$ & $64.3$ & $57.2$ & $63.4$\\
    & OSR & $54.9$  & $31.9$ & $47.5$ & $52.0$ & $53.6$ & $56.4$  & $45.4$ & $45.8$ & $44.5$ & $45.9$\\
   \hline
     \end{tabular}\label{table:otb}
\end{table*}

\section{Conclusion} \label{Conc}
In this paper, we propose a particle filter redetection tracker with correlation filters (CFPFT) to achieve an effective and robust performance on the test benchmark. The redetection mechanism  plays an important role in the visual object tracking process when the tracker losts its object. It can effectively re-locate the object and improve the tracking performance by extensive particle resampling that can provide more candidates.
Besides, we give a simple scale evaluation mechanism  that shows the effectiveness on the sequence with scale change.
The extensive experimental results shows the competitiveness of our CFPFT tracker compared with $37$ trackers, which are widely used in the performance evaluation of tracking algorithm.
The analysis of the experimental results with different attributes demonstrates the better ability of our tracker.

\section*{Acknowledgment}
This research was supported by the National Natural Science Foundation of China (Grant No.61672183),
by the Shenzhen Research Council (Grant Nos. JCYJ20170413104556946, JCYJ20160406161948211, JCYJ20160226201453085), and by the Natural Science Foundation of Guangdong Province (Grant No. 2015A030313544).

\bibliographystyle{elsarticle-num}
\bibliography{mybibfile}
\end{document}